\documentclass{article}

\PassOptionsToPackage{numbers,sort&compress}{natbib}
\usepackage[preprint]{neurips_2026}
\usepackage{amsmath,amssymb}
\usepackage{wrapfig}
\newcommand{\R}{\mathbb{R}}

\usepackage[utf8]{inputenc} %
\usepackage[T1]{fontenc}    %
\usepackage{hyperref}       %
\usepackage{url}            %
\usepackage{booktabs}       %
\usepackage{amsfonts}       %
\usepackage{nicefrac}       %
\usepackage{microtype}      %
\usepackage{xcolor}         %

\usepackage{microtype}
\usepackage{graphicx}
\usepackage{subcaption}
\usepackage{booktabs} %
\usepackage{subcaption}
\usepackage{booktabs}
\usepackage[dvipsnames]{xcolor}
\usepackage{wrapfig}
\usepackage{amsmath}
\usepackage{amssymb}
\usepackage{mathtools}
\usepackage{amsthm}
\usepackage[dvipsnames]{xcolor}

\definecolor{LightBlue}{RGB}{80,160,255}
\definecolor{CitePurple}{RGB}{120,70,200}
\usepackage{hyperref}

\hypersetup{
  colorlinks=true,
  citecolor=CitePurple,
  linkcolor=LightBlue,
  urlcolor=LightBlue
}

\title{Branched Optimal Transport Amortization}

\author{%
  Semyon Semenov$^1$ \And
  Viktor Kovalchuk$^1$ \And
  Meir Roketlishvili$^1$ \And
  Albert Baichorov$^1$ \AND
  Fakhri Karray$^1$ \And
  Martin Tak\'a\v{c}$^1$ \And
  Arip Asadulaev$^1$
}

\begin{document}

\maketitle
\footnotetext[1]{MBZUAI. Correspondence to: Semyon Semenov $<$\texttt{semen.semenov@mbzuai.ac.ae}$>$.}
\setcounter{footnote}{1}

\begin{abstract}
  Methods of Branched Optimal Transport (BOT) mimic the economy and efficiency of natural tree-like structures, such as those found in rivers and biological systems. These methods are widely applicable for designing efficient networks in society, from river basins and blood vessels to mail and gas distribution systems. However, they remain understudied in the context of designing deep generative models, particularly at a large scale. Standard continuous-time generative models, such as the flow matching approach, fail to capture the inherent hierarchical and branching patterns present in real-world data. Current models provide no mechanism for flows to merge or share pathways to minimize total transport cost. Inspired by the "economy of scale" principle in BOT, we introduce a novel, scalable branched flow-matching algorithm designed to solve the branched optimal transport problem in high dimensions. Our method adapts the Benamou-Brenier continuous-time optimal transport formulation to learn branched generative flows. These flows allow probability mass to aggregate along common pathways before branching out to diverse targets. Parametrized by neural networks, our method effectively learns complex branched generative processes. We demonstrate its effectiveness on challenging high-dimensional tasks in biology and image generation.
\end{abstract}

\section{Introduction}
\label{sec:intro}
\begin{wrapfigure}{r}{0.5\textwidth}
    \centering
    \includegraphics[width=0.38\linewidth]{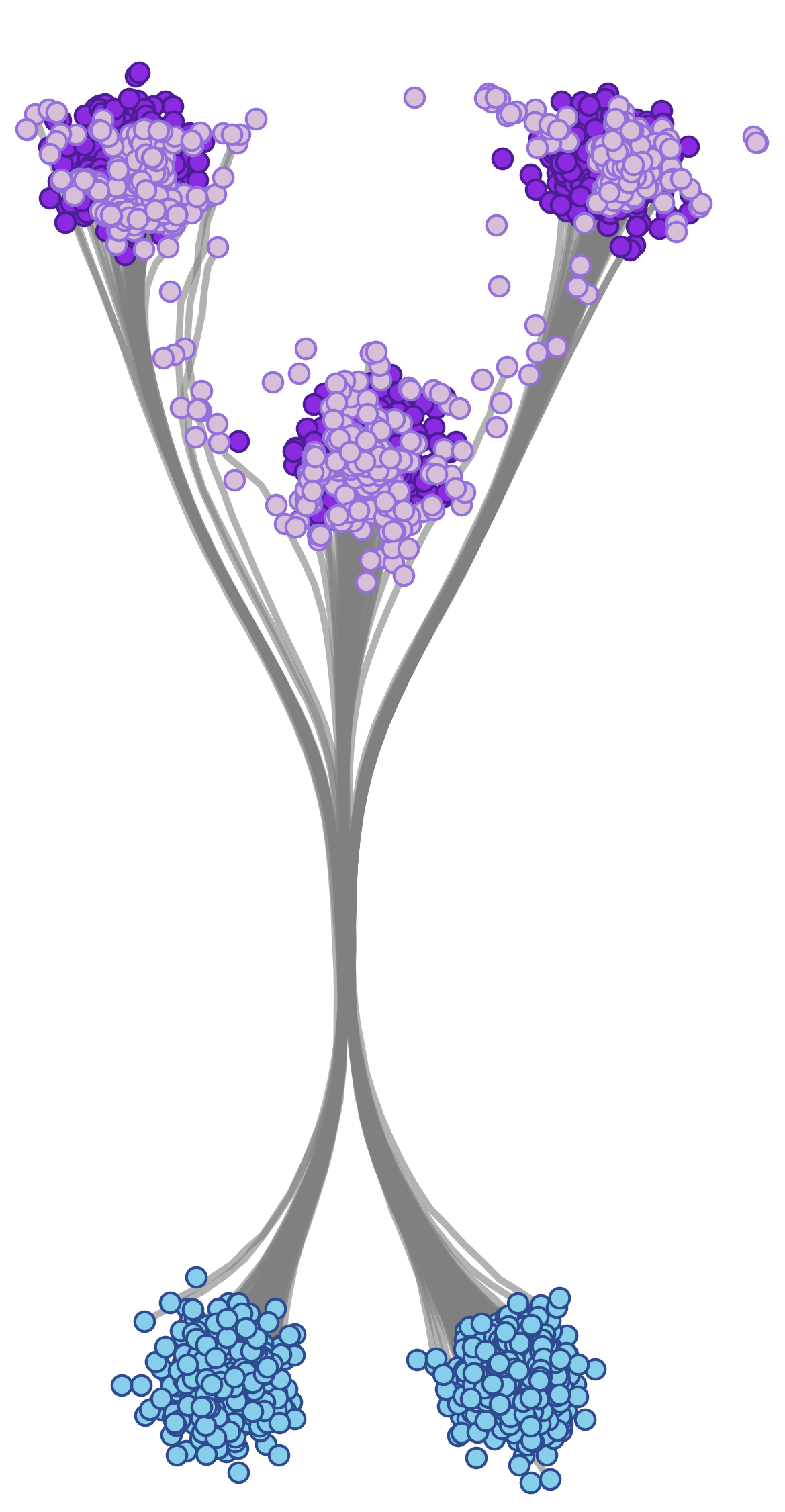}
    \includegraphics[width=0.55\linewidth]{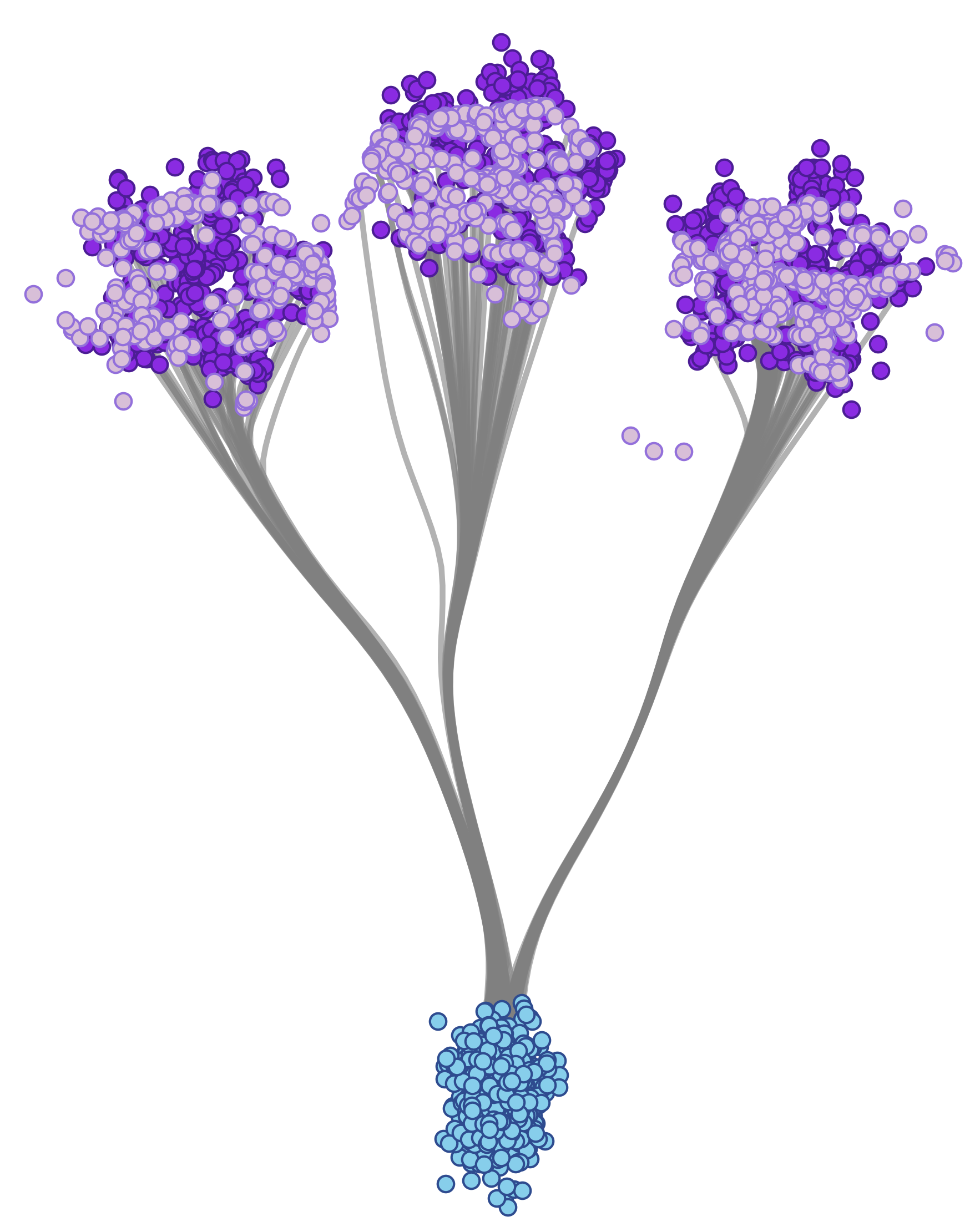}
    \caption{BOTA Gaussian results for $\alpha = 0.5$.}
    \vspace{-5mm}
\end{wrapfigure}
Deep generative modeling has advanced rapidly, with diffusion models and simulation-free flow–based methods (e.g., flow matching and rectified flows) now supporting state‑of‑the‑art synthesis, fast sampling, and scalable training. These approaches learn vector fields (or stochastic dynamics) that transport a simple base distribution to complex data distributions, and many of them explicitly encourage \textit{straight} transport trajectories between coupled samples—mirroring the \emph{displacement interpolation} paradigm in classical optimal transport (OT)~\citep{ho2020denoising, lipman2022flow, tong2023improving, liu2022flow}.

However, straight-line, independent motions of mass are often a poor fit for data with hierarchical or multi-modal organization. ImageNet, for example, is built on the WordNet taxonomy: “golden retriever” and “labrador” are siblings under “dog,” which is nested under broader synsets such as “canine” and “mammal.” It is natural to expect that the generative process should \emph{share substantial computation} along the path common to “dogs” before specializing to particular breeds. Such structures arise even more naturally and meaningfully in biological processes, which can also be observed as empirical distributions. Modeling such shared structure is difficult.

\textbf{Branched optimal transport} generalizes classical OT by introducing economies of scale: moving two units of mass together along the same route is cheaper than moving them separately. Formally, BOT minimizes energies in which the cost along an edge is \textit{subadditive} in the mass (often proportional to $m^\alpha$ with $0 < \alpha < 1$), which causes optimal solutions to \textit{coalesce} into \textbf{tree- or network-like structures} rather than independent straight paths. This mechanism closely matches how efficient networks form in nature and society, from river basins to mail and gas distribution systems~\citep{buttazzo2003optimal, xia2003optimal, bernot2005traffic}.

The BOT perspective is powerful for generative modeling because it encodes \textbf{shared prefixes of computation}: samples that belong to nearby leaves in a taxonomy naturally travel together for most of the journey, branching only when necessary. In contrast with Euclidean quadratic-cost OT—where displacement interpolation moves particles independently along straight segments defined by a Brenier map—BOT explicitly rewards branching, aligning the transport with hierarchical data.

Mathematically, BOT has a mature foundation: it has been studied via discrete “irrigation” models, continuum formulations, and relaxations~\cite{bernot2009optimal}. Beyond the modeling appeal, BOT solutions satisfy rich structural properties (e.g., optimal junctions have bounded degree), reflecting the network nature of the minimizers~\cite{santambrogio2015optimal, bernot2009optimal, oudet2011modica}. Despite its appeal, \textbf{BOT is algorithmically challenging}. Even in planar settings with finitely many sources/sinks, optimizing BOT networks is \textit{NP-hard}, and require solving large PDE systems, which becomes prohibitive at modern data scales. This has kept BOT largely confined to small or low-dimensional instances.

At the same time, contemporary generative models operate at unprecedented scale. Flow matching and rectified-flow variants are trained on web-scale datasets and large backbones, and they benefit from transport formulations that are differentiable end-to-end and hardware-accelerated. A \textbf{scalable BOT solver} would let us \textit{replace independent straight-path assumptions} by \emph{economy-of-scale transport} that respects data hierarchies practical training pipelines.

\textbf{Neural parameterizations are a natural fit.} In standard OT, neural solvers already amortize transport maps or potentials and integrate cleanly with deep learning. Extending this spirit to BOT promises (i) \underline{\textit{amortized inference}} of branched routes that can be reused across batches and tasks,

\textbf{Contribution}: We introduce \textbf{Branched Optimal Transport Amortization (BOTA)}, a continuous and scalable solver for branched optimal transport. Our core contribution is a \emph{novel optimization problem} that we adapt into a generative algorithm, which is efficiently parameterized by neural networks using a \underline{flow-matching objective}. We demonstrate the effectiveness of our method on challenging biological and image generation tasks, showing that it successfully learns meaningful hierarchical generative processes that reflect the underlying structure of the data.
\vspace{-3mm}
\section{Background}
\label{sec:premilinaries}
\vspace{-3mm}
\textbf{Notation.}
Let $\Omega \subset \mathbb{R}^d$ be a compact and convex domain. We denote by $\mathcal{P}_2(\Omega)$ the space of probability measures on $\Omega$ with a finite second moment. When a measure $\mu$ admits a density relative to the Lebesgue measure, we denote its density by $\rho$ (i.e., $d\mu(x) = d\rho(x)dx$). Vectors are column vectors, $\|\cdot\|$ is the standard Euclidean norm, and $\nabla\!\cdot$ is the divergence operator and  $\|\cdot\|_F$ is the Frobenius norm.

\paragraph{Continuous Normalizing Flow}
\label{sec:cnf}
A Continuous Normalizing Flow (CNF) \cite{chen2018neural} is a generative model that defines a probability density path through a Neural Ordinary Differential Equation (ODE):
\begin{equation}
  \label{eq:cnf-ode}
  \frac{d}{dt} x_t = v_\theta(x_t,t), \qquad x_{t=0} \sim \mu_0.
\end{equation}
Let $\Phi_t$ be the flow map associated with this ODE, which transports a particle from its initial condition at time $0$ to its location at time $t$. The pushforward density $\rho_t = (\Phi_t)_\# \rho_0$ evolving under this dynamics \emph{necessarily} satisfies the \emph{continuity equation} ~\eqref{eq:continuity} with the parameterized velocity field $v_\theta$. Where continuity equation encodes mass conservation:
\begin{equation}
  \partial_t \rho_t + \nabla \cdot (\rho_t v_t) = 0 \nonumber
  \quad \text{on } \Omega \times (0,1), \quad
  \rho_{t=0}=\rho_0,\quad \rho_{t=1}=\rho_1.
  \label{eq:continuity}
\end{equation}
A key result is the change of variables formula, which describes how the log density evolves:
\begin{equation}
  \label{eq:cnf-logdet}
  \frac{d}{dt}\,\log \rho_t(x_t) = -\,\nabla\!\cdot v_\theta(x_t,t). 
\end{equation}
This allows for a likelihood calculation by integrating this quantity over time. Training can be done by directly maximizing likelihood (integrating \eqref{eq:cnf-logdet}).%
\paragraph{Flow Matching.} The core idea of Flow Matching (FM)\citep{lipman2022flow, tong2023improving} is to train a CNF by directly regressing its velocity field $v_\theta$ toward a target vector field $u_t$ that generates a desired probability path. The Flow Matching objective is: $\mathcal{L}_{\text{FM}}(\theta)
  = \mathbb{E}_{t\sim \mathcal{U}[0,1],\, x\sim \rho_t}\!\left[\|v_\theta(x,t)-u_t(x)\|^2\right]$.

A critical challenge is that sampling $x \sim \rho_t$ from the marginal path at arbitrary times is typically intractable. Conditional Flow Matching (CFM) \cite{lipman2022flow} provides a solution by constructing the marginal path as a mixture of simpler and tractable conditional paths. Let $z$ be a conditioning variable with distribution $q(z)$. We define the marginal path as follows: $\rho_t(x) = \int \rho_t(x\mid z)  q(z)\,dz$, where each conditional path $\rho_t(x|z)$ is generated by a corresponding conditional vector field $u_t(x\mid z)$. The marginal field $u_t(x)$ that generates $\rho_t$ is then given by:
\begin{equation}
  \label{eq:marginal-field}
  u_t(x) = \mathbb{E}_{z\sim q}\!\left[ \frac{\rho_t(x\mid z)}{\rho_t(x)}\,u_t(x\mid z) \right],
\end{equation}
where $q(z\mid x)$ is the posterior. The key theorem \cite{lipman2022flow} is to minimize the \emph{Conditional Flow Matching}:
\begin{equation}
  \label{eq:cfm}
  \mathcal{L}_{\text{CFM}}(\theta)
  = \mathbb{E}_{t,\, z\sim q,\, x\sim \rho_t(\cdot\mid z)}\!\left[\|v_\theta(x,t)-u_t(x\mid z)\|^2\right]
\end{equation}
with $t\sim\mathcal{U}[0,1]$ yields the same gradient for $\theta$ as minimizing the intractable $\mathcal{L}_{\text{FM}}(\theta)$ in \eqref{eq:cfm}. This makes CFM a practical objective, as it only requires sampling from conditional paths $\rho_t(x|z)$ and knowing their closed-form drifts $u_t(x|z)$.

The flexibility of CFM lies in the choice of conditional paths. The coupling $q(z)$ is the independent joint distribution $q(x_0)q(x_1)$, so $z=(x_0, x_1)$. A common conditional path is a Gaussian bridge: $\rho_t(x\mid z) = \mathcal{N}(x \mid \mu_t, \sigma^2 I)$, where $\mu_t = (1-t)x_0 + t x_1$ is a linear interpolation. The simple constant drift that generates this path is $u_t(x\mid z)=x_1-x_0$. 

\textbf{Monge--Kantorovich Optimal Transport (OT)}
\label{sec:static_ot}
seeks a way to morph one probability distribution with minimal effort, as quantified by a cost function $c(x, y)$. The Monge formulation seeks a deterministic map $T: \Omega \to \Omega$ that pushes $\mu_0$ to $\mu_1$ and minimizes the total cost $\int c(x, T(x)) d\mu_0(x)$ \citep{villani2008optimal}. This problem can be ill-posed, its relaxation, the Kantorovich problem, searches over \emph{couplings} (joint distributions) $\gamma \in \Gamma(\mu_0,\mu_1)$ with marginals $\mu_0$ and $\mu_1$: 
\begin{equation}
    \inf_{\gamma \in \Gamma(\mu_0,\mu_1)}
  \int_{\Omega \times \Omega} c(x,y)\, d\gamma(x,y).
  \label{eq:kant}
\end{equation}
For cost $c(x,y)=\|x-y\|^2$, the square root of the solution is the Wasserstein-2 distance. The flow matching approach discussed in the section above can be connected with the flow matching to create the optimal transport CFM (OT-CFM) \citep{tong2023improving} method uses an optimal coupling to define conditionals. Here, $z=(x_0, x_1)$ is sampled from an OT plan $\gamma$ between $\mu_0$ and $\mu_1$ \ref{eq:kant}. %
In practice, the OT plan $\gamma$ is efficiently approximated using mini-batch OT by Sinkhorn algorithm~\cite{cuturi2013sinkhorn}.

\textbf{Benamou--Brenier OT} 
frames transportation as a continuous-time problem. For flow $\Phi_t$, we have a pushforward density $\rho_t = (\Phi_t)_\# \rho_0$ evolving under dynamics that satisfies \emph{continuity equation} \ref{eq:continuity} with the velocity field $v$.
The Benamou--Brenier theorem states that the squared Wasserstein distance equals the minimal \emph{kinetic energy} of such a flow:
\begin{equation}
  \label{eq:bb}
  \inf_{\rho_t,\, v_t}
  \left\{ \int_0^1 \!\! \int_{\Omega} \|v_t(x)\|^2 \,\rho_t(x)\, dx\, dt \quad \text{subject to \eqref{eq:continuity}} \right\}.
\end{equation}
The minimizers of this problem are constant-speed geodesics in the Wasserstein space. When an optimal transport map $T$ exists for the static problem (e.g., when $\mu_0$ is absolutely continuous), the geodesic is given by \emph{interpolation}: $\rho_t = ((1-t)\mathrm{Id}+tT)_\# \rho_0$. The corresponding velocity field is constant along the trajectories, $v_t(x_t) = T(x_0)-x_0$, and the particle dynamics is linear: $x_t = (1 - t)x_0 + tT(x_0)$.

\section{Branched Benamou–Brenier OT}
\label{subsec:branched_ot_bb}
Branched Benamou–Brenier formulation studied by~\cite{brasco2011benamou}, which is also equivalent to the formulation of~\cite{xia2003optimal}. The idea is to consider the classical Benamou–Brenier optimal transport problem \ref{eq:bb}, but with a \emph{concave} cost function. The boundary conditions remain the same as in standard optimal transport: $\rho(0,\cdot)=\mu_0$ and $\rho(1,\cdot)=\mu_1$. With $\alpha \in (0,1)$ as the branching parameter, we define the cost that measures the \textit{work} of moving a mass $m$ over a distance $\ell$, simply written as $m^\alpha \ell$.  If $\alpha=1$, the cost becomes linear in the mass ($m^1 \ell$), recovering the classical OT setting \ref{sec:static_ot}. By identifying the length $\ell$ with the velocity $v$ and the mass $m$ with the time-dependent probability density $\rho_t$, the dynamic branched optimal transport can be formulated as:
\begin{equation}
  \label{eq:branched_bb}
  B_{\alpha}(\mu_0,\mu_1)
  \;=\;
  \inf_{\substack{\rho(t,\cdot),\,v(t,\cdot)}}
  \int_{0}^{1}
  \sum_{i\in I_t} \lvert v_{t,i}\rvert\,{\textcolor{Orange}{(\rho_{t,i})^\alpha}}
  \,dt.
\end{equation}
What the reader can notice is that for the cost \eqref{eq:branched_bb} need to be finite, the measure $\rho(t,\cdot)$ must be \emph{purely atomic} for (almost every) time $t$! This means that the mass is concentrated at a points $\{x_{t,i}\}_{i\in I_t}$:
    \[
    \rho(t,\cdot) = \sum_{i\in I_t} \rho_{t,i}\,\delta_{x_{t,i}}, \quad \text{where } \rho_{t,i} = {\textcolor{Green}{\rho(t, \{x_{t,i}\})}}.
    \]
The associated momentum is then $q = \rho v = \sum_{i\in I_t} v_{t,i}\,\rho_{t,i}\,\delta_{x_{t,i}}$, where $v_{t,i} = v(t, x_{t,i})$ is the velocity of the atom $x_{t,i}$. This formulation is computationally intensive, which motivates our development of a more efficient relaxation.  

\section{Branched Benamou Brenier Neural OT}
\label{sec:BBBOT}
To build a Benamou Brenier style Branced OT flows we need to develop a flow algorithms that meet a few constraints:
\begin{itemize}
    \item {\textcolor{Blue}{The boundary constraint}}: we need a flow whose starting and end points are equal to the given distributions: $\rho(0,\cdot)=\mu_0$ $\rho(1,\cdot)=\mu_1$.
    \item {\textcolor{Orange}{Mass-dependent cost}}: We need to have a time-dependent density function that we are using as part of the cost function. 
    \item {\textcolor{Green}{Atomic mass constraint}}: Our density function must be atomic almost every time $t$, so the mass needs to be concentrated in a countable set of points. 
\end{itemize}
To meet these constraints, it is convenient to draw an analogy between branched optimal transport and continuous normalizing flows. Having a continued vector field, we can derive a time continuous {\textcolor{Orange}{probability density}} function that plays a central role in the formulation of branched optimal transport. Also, CNFs by design meet the {\textcolor{Blue}{the boundary constraint}} \ref{sec:cnf}. So, first, let us parameterize the velocity field by a continuous normalizing flow.
\[
\dot x_t=v_\theta\!\big(x_t,t\big),\qquad t\in[t_0,t_1],
\]
whose density \(p_\theta(x,t)\) evolves under the instantaneous change–of–variables, CNF formula:
\begin{equation}
\label{eq:cnf-cov}
\log p_\theta\!\left(x_{t_1},t_1\right)
=
\log p_\theta\!\left(x_{t_0},t_0\right)
-
\int_{t_0}^{t_1}\!\! \operatorname{tr}\!\big(\partial_x v_\theta(x_t,t)\big)\,dt .
\end{equation}
This enforces the continuity equation by construction, making CNFs natural for dynamic OT objectives. To capture branched transport, we concavify the kinetic action by the density, mirroring the BOT cost \(m^\alpha \ell\) with \(\alpha\in(0,1)\). Our learning objective over terminal samples is
\begin{equation}
\begin{aligned}
\mathcal{L}^\alpha(\theta) &= 
\int_{t_0}^{t_1} \!\! \int_{\Omega} \underbrace{
    \big\| v_\theta(x,t) \big\|
    {\textcolor{Orange}{p_\theta(x,t)^{\alpha}}}
    \, dx \, dt}_{\textcolor{Black}{\text{branched cost}}}  - \lambda \underbrace{ 
    \mathbb{E}_{x_{t_1} \sim \mu_1} \!\big[ \log p_\theta(x_{t_1}, t_1) \big]
    }_{\textcolor{Blue}{\text{match data at } t_1}}.
\label{eq:bbot_cnf}
\end{aligned}
\end{equation}

The second term is the CNF maximum-likelihood fit to \(\mu_1\) at \(t_1\).
The first term is a BOT-style dynamic action: it penalizes kinetic effort \(\|v_\theta\|\) but discounts it by \(p_\theta^\alpha\) with \(\alpha<1\). Our method is a concave reweighting that implements the BOT principle \textit{move together, then split}, while keeping the CNF machinery and training recipe. 

A challenge in implementing the BOT objective \eqref{eq:bbot_cnf} within a CNF framework is the \emph{atomic mass constraint}. The theoretical BOT formulation requires the evolving measure $\rho(t,\cdot)$ to be a sum of discrete point masses for the cost to be finite. However, CNFs naturally define a \emph{continuous} density field $p_\theta(x,t)$, making these two concepts fundamentally incompatible for optimization.

To bridge this gap, we introduce a differentiable relaxation of the atomic constraint, which we term the \textit{soft-atomic cost}. In a computational setting, we cannot handle continuous measures or an infinite number of atoms. We make two approximations. Instead of forcing the continuous density to collapse into discrete points, this objective encourages the density to \emph{cluster} around a predefined set of $K$ proxy atoms. This is achieved by softly assigning the mass of particles from the flow to these centers based on proximity.

At any given time t, for a batch of particles {$x_t$} sampled from the flow density $p_\theta(\cdot,t)$, we define the soft mass $m_k(t)$ and the average velocity $v_k(t)$ associated with center $c_k$ and where $w_k(x)$ is a soft-assignment weight, typically a Gaussian kernel that measures the influence of center $c_k$ on a particle at position $x$:
\begin{align}
w_k(x) = \frac{\exp(-\frac{\|x-c_k\|^2}{2 \sigma^2})}{\sum_{j=1}^K \exp(-\frac{\|x-c_j\|^2}{2 \sigma^2})},\quad
m_k(t) = \mathbb{E}_{x_t \sim p_\theta(\cdot, t)} \left[ w_k(x_t) \right],
\end{align}
\begin{equation}
\label{eq:soft_velocity}
v_k(t) = \frac{\mathbb{E}_{x_t \sim p_\theta(\cdot, t)} \left[ v_\theta(x_t, t) w_k(x_t) \right]}{m_k(t)}.
\end{equation}
The instantaneous soft-atomic cost is then formulated as a differentiable proxy to the theoretical BOT cost. By integrating soft cost over time, we reformulate the learning objective as: 
\begin{equation} 
\label{eq:soft-atomic-loss} 
\mathcal{L}^\alpha(\theta) =\int_{t_0}^{t_1}\sum_{k=1}^{K} \underbrace{v_k(t) {\textcolor{Orange}{m_k(t)^{\alpha}}} dt
}_{\text{\textcolor{Green}{soft branched cost}}} - \lambda\underbrace{ \mathbb{E}_{x_{t_1}\sim \mu_1}\!\big[\log p_\theta(x_{t_1},t_1)\big] }_{\textcolor{Blue}{\text{match data at} t_1}}.
\end{equation}

\textbf{Lemma 1} (Differentiable Approximation of the Instantaneous Branched Transport Cost)
\textit{Let $\hat{\rho}_t = \frac{1}{B} \sum_{j=1}^B \delta_{x_j(t)}$ be an empirical probability measure representing a distribution at time $t$ with particles at positions $\{x_j\}_{j=1}^B$ moving with velocities $\{v_j\}_{j=1}^B$. Let $\{c_k\}_{k=1}^K$ be a set of fixed centers in $\mathbb{R}^d$. The function soft atomic cost, defined as:
$$ C_{\text{soft}}( \{x_j\}, \{v_j\} ) = \sum_{k=1}^K |v_k| m_k^\alpha, $$
where for each center $c_k$:
\begin{align*}
    m_k = \frac{1}{B} \sum_{j=1}^B w_{jk} \quad (\text{Soft Mass})\qquad 
    |v_k| = \frac{\sum_{j=1}^B w_{jk} |v_j|}{\sum_{j=1}^B w_{jk}} \quad (\text{Soft Average Speed})
\end{align*}
with weights $w_{jk} = \text{softmax}(-||x_j - c_k||^2 / 2\sigma^2)_k$, is a differentiable approximation of the true instantaneous branched transport cost functional for the discretized measure. In the limit as the kernel width $\sigma \to 0$, this cost converges to the cost of a hard assignment of particles to their nearest centers.} Please see proofs in Appendix. 

\section{Branched Flow Matching via BOT Amortization}

The primary challenge of branched optimal transport, as formulated in Sec.\ref{subsec:branched_ot_bb}, is the difficulty of directly optimizing the branched Benamou-Brenier objective \eqref{eq:bbot_cnf} over the space of continuous-time neural velocity fields. The objective is non-trivial, and the "atomic" constraint is inherently discrete.

To bridge this, we introduce a practical two-stage method, Branched Optimal Transport Amortization (BOTA) also we call it \textbf{Branched Flow Matching}, which first solves a discrete-time version of the our proposed BOT soft-atomic relaxation for a fixed batch and then amortizes this solution into a continuous-time neural network using a flow matching objective. This approach is analogous to standard OT-CFM \eqref{eq:bbot_cnf}, but we replace the simple, straight-line (Euclidean) optimal transport plan Sec. \ref{sec:static_ot} with a more complex, branched transport plan.

\subsection{Discrete Branched Transport Solver}

First, we compute an exemplar branched flow for a single, fixed batch of $N$ source samples $\{x_0^i\}_{i=1}^N \sim \mu_0$ and $N$ target samples $\{x_1^i\}_{i=1}^N \sim \mu_1$. We discretize the time interval $[0, 1]$ into $T$ steps of size $\Delta t = 1/T$. Our goal is to find an optimal discrete velocity field $V \in \R^{T \times N \times D}$ that transports the batch $\{x_0^i\}$ to $\{x_1^i\}$ while minimizing a discrete analogue of the branched transport cost. The particle trajectories $\{X_k\}_{k=0}^T$, where $X_k = \{x_k^i\}_{i=1}^N$, are computed via standard Euler integration:
$$
x_{k+1}^i = x_k^i + V_{k,i} \cdot \Delta t, \quad \text{with } x_0^i \text{ given.}
$$
Here, $V_{k,i} \in \R^D$ is the velocity of particle $i$ at discrete time step $k$.

We optimize $V$ by minimizing a loss function that combines a terminal matching cost with the time-integrated soft-atomic energy:
$$
\mathcal{L}_{\text{discrete}}(V) = \sum_{k=0}^{T-1} C_{\text{soft}}(X_k, V_k) \Delta t + \lambda \frac{1}{N} \sum_{i=1}^N \|x_T^i - x_1^i\|^2. 
$$
The first term is a standard mean-squared error that forces the final positions $x_T^i$ to match the target positions $x_1^i$. The second term is the discrete integral of the instantaneous branched transport cost, $C_{\text{soft}}$, which is defined precisely in Lemma 1 (Eq. 14) as:
$$
C_{\text{soft}}(X_k, V_k) = \sum_{j=1}^{K} |v_j|_k (m_j)_k^{\alpha},
$$
where $(m_j)_k$ and $(|v_j|)_k$ are the soft mass and soft average speed, respectively, associated with the $j$-th proxy atom $c_j$ at time step $k$.

It is important to note that our terminal matching term, $\lambda \frac{1}{N} \sum_{i=1}^N \|x_T^i - x_1^i\|^2$, is a simple Mean Squared Error (MSE). This is a deliberate choice for computational efficiency. Since this discrete solver operates on a fixed, finite batch, we are computing a deterministic, point-to-point map for these specific $N$ samples. We are not training a generative model to match the full distribution $\mu_1$ at this stage. Therefore, a direct MSE is the most appropriate and efficient loss to enforce the terminal constraint, in contrast to a heavier probabilistic loss (like a CNF log-likelihood) which would be unnecessary for this exemplar-finding step. By solving this optimization problem (e.g., via gradient descent on $V$), we obtain an optimal discrete velocity field $V^* \in \R^{T \times N \times D}$ and its corresponding optimal trajectory $X^* \in \R^{(T+1) \times N \times D}$. This pair $(X^*, V^*)$ represents a single, discrete branched transport path for the given batch.

\subsection{Amortization via Flow Matching}

The discrete solution $(X^*, V^*)$ is only valid for the specific batch it was trained on. To create a generative model, we must learn a continuous-time neural velocity field $v_{\theta}(x, t)$ that can generalize this branched behavior. We achieve this by "distilling" the discrete solution into $v_{\theta}$ using a flow matching objective. This step reframes the problem as a standard supervised regression task. We treat the discrete solution $(X^*, V^*)$ as the ground truth. We construct a continuous-time target vector field, $u_t(x)$, by linearly interpolating the discrete solution. For continuous time $t \in [0, 1]$, we find its corresponding discrete time index $k = \lfloor t / \Delta t \rfloor$ (clamped to $[0, T-1]$) and the interpolation weight $\beta = (t - k \Delta t) / \Delta t \in [0, 1]$. For each particle $i$, the target position $x_t^i$ and target velocity $u_t(x_t^i)$ at time $t$ are defined by linear interpolation:
\begin{align*}
    x_t^i = (1 - \beta) x_k^{*,i} + \beta x_{k+1}^{*,i}, \qquad
    u_t(x_t^i) = (1 - \beta) V_{k,i}^* + \beta V_{k',i}^*,
\end{align*}
where $k' = \min(k+1, T-1)$ is the next velocity index.

The Branched Flow Matching (BFM) objective is then to regress the neural network $v_{\theta}$ onto this continuous, interpolated target field $u_t$:
$$
\mathcal{L}_{\text{BFM}}(\theta) = \mathbb{E}_{i \sim \mathcal{U}[1,N], t \sim \mathcal{U}[0,1]} \left[ \left\| v_{\theta}(x_t^i, t) - u_t(x_t^i) \right\|^2 \right].
$$
By minimizing this objective, the network $v_{\theta}$ learns to approximate the complex, branched velocity field derived from the discrete BOT solver. Once trained, $v_{\theta}$ can be used as a generative model to transport new samples from $\mu_0$ to $\mu_1$ by solving the continuous-time ODE $\dot{x} = v_{\theta}(x, t)$, effectively amortizing the expensive discrete optimization.

\begin{figure*}[t!]
\centering

\begin{subfigure}[b]{0.28\linewidth}
    \centering
    \includegraphics[height=0.28\textheight]{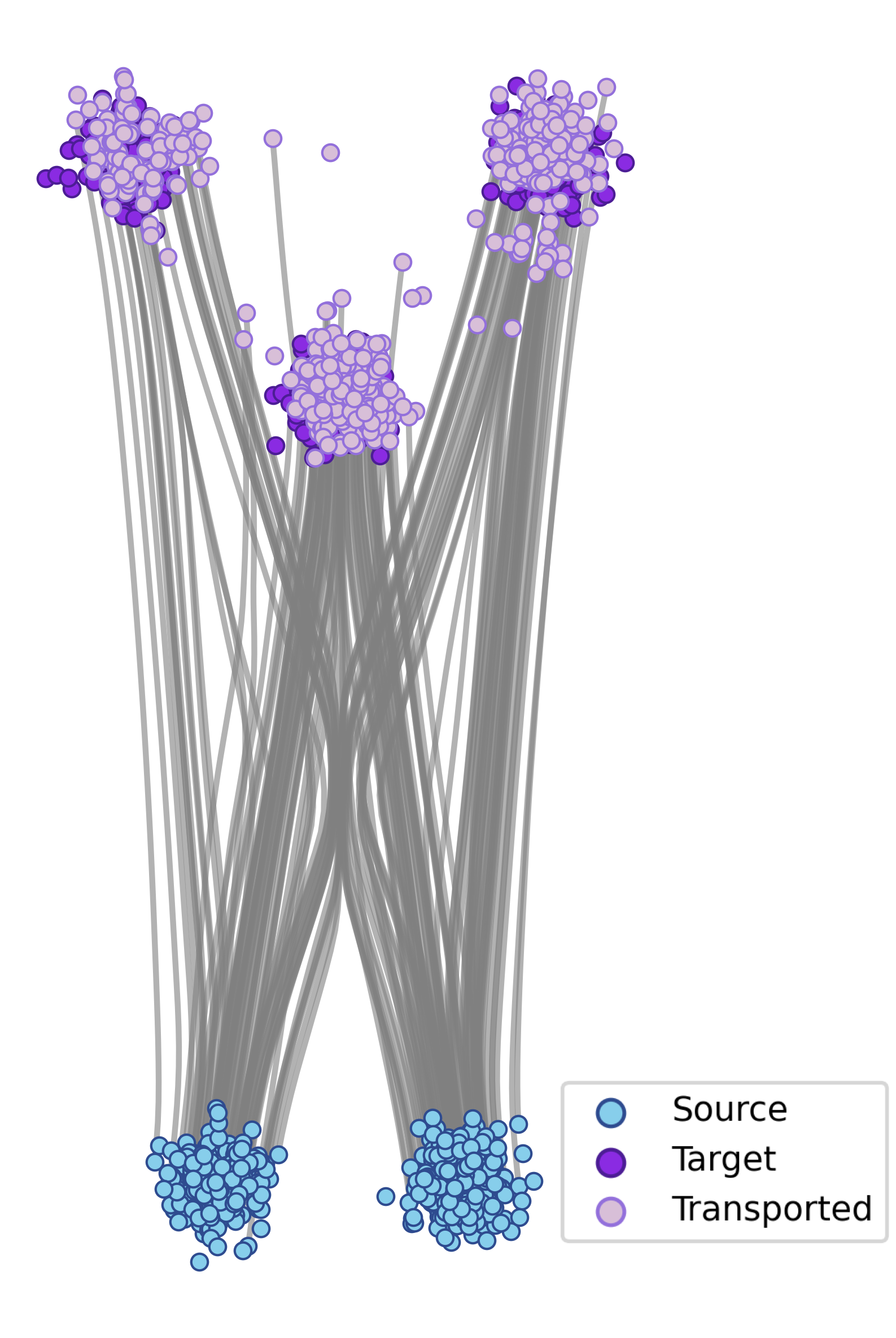}
    \caption{3  (Flow Matching)}
\end{subfigure}
\begin{subfigure}[b]{0.28\linewidth}
    \centering
    \includegraphics[height=0.28\textheight]{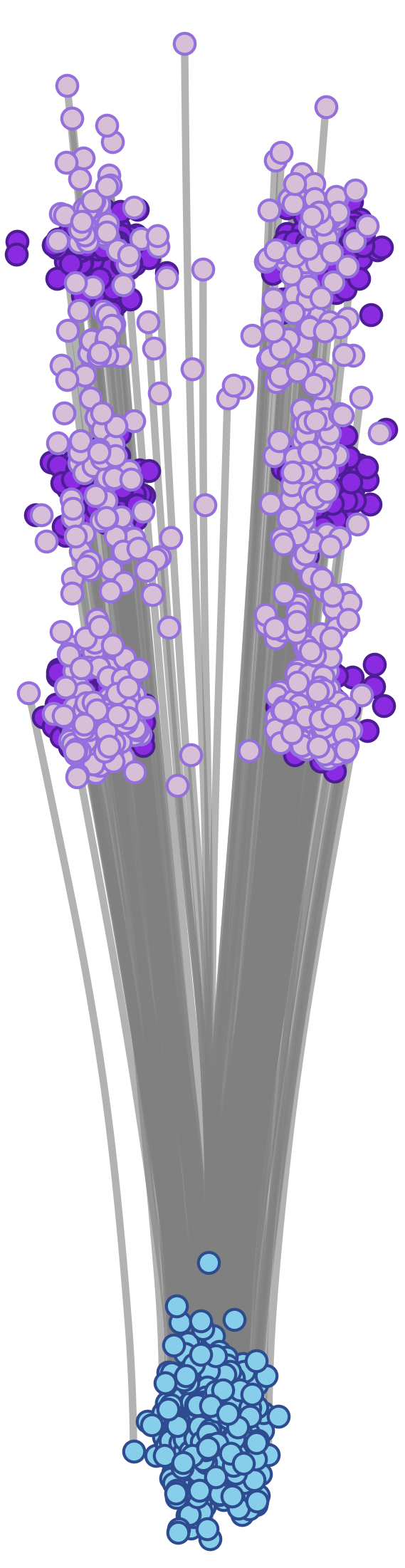}
    \caption{6  (Flow Matching)}
\end{subfigure}
\begin{subfigure}[b]{0.28\linewidth}
    \centering
    \includegraphics[height=0.28\textheight]{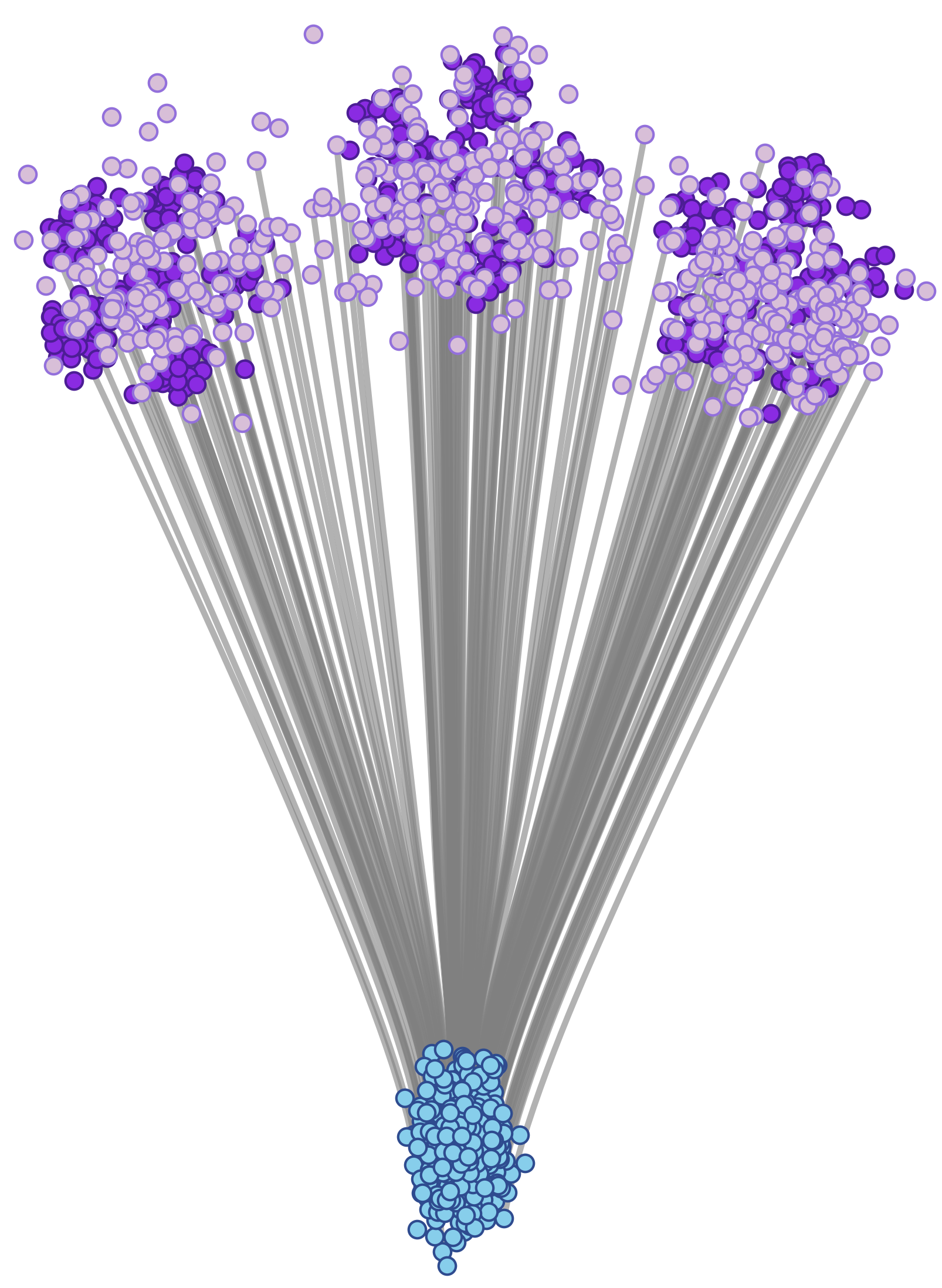}
    \caption{18  (Flow Matching)}
\end{subfigure}

\vspace{3mm}

\begin{subfigure}[b]{0.28\linewidth}
    \centering
    \includegraphics[height=0.28\textheight]{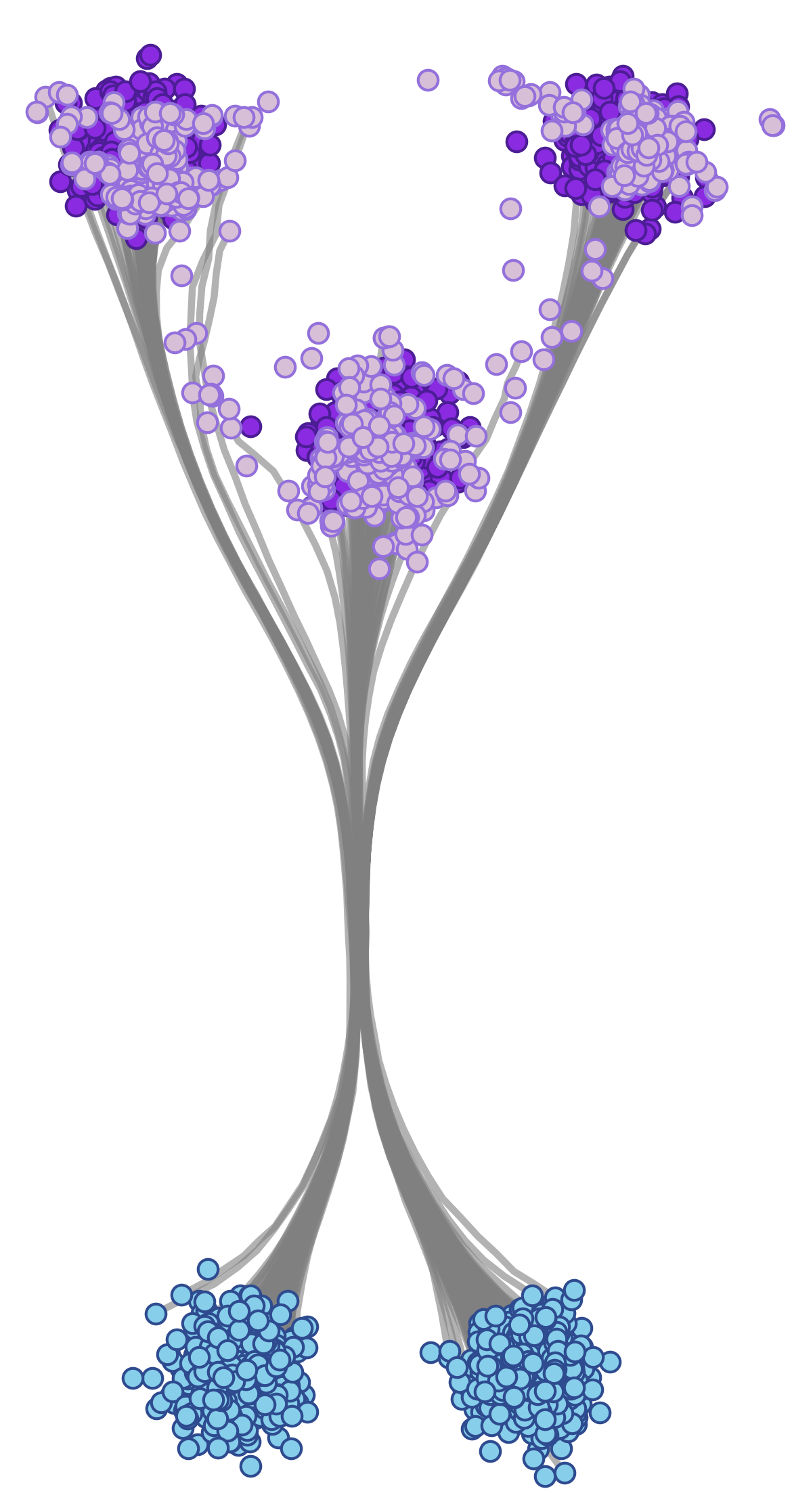}
    \caption{3  (BOTA)}
\end{subfigure}
\begin{subfigure}[b]{0.28\linewidth}
    \centering
    \includegraphics[height=0.28\textheight]{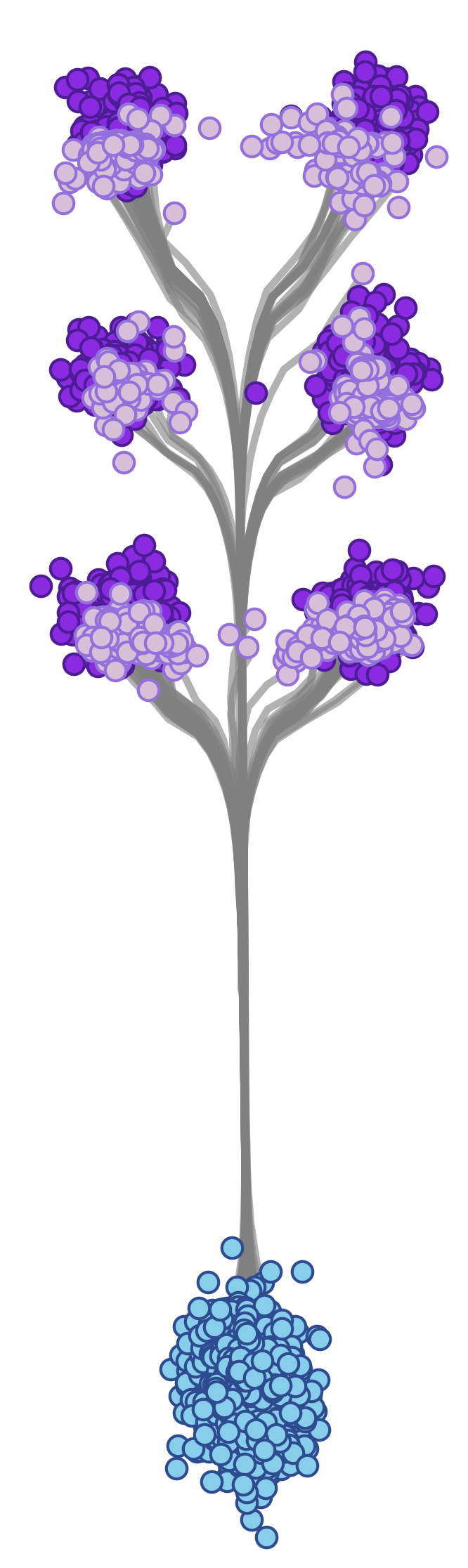}
    \caption{6  (BOTA)}
\end{subfigure}
\begin{subfigure}[b]{0.28\linewidth}
    \centering
    \includegraphics[height=0.28\textheight]{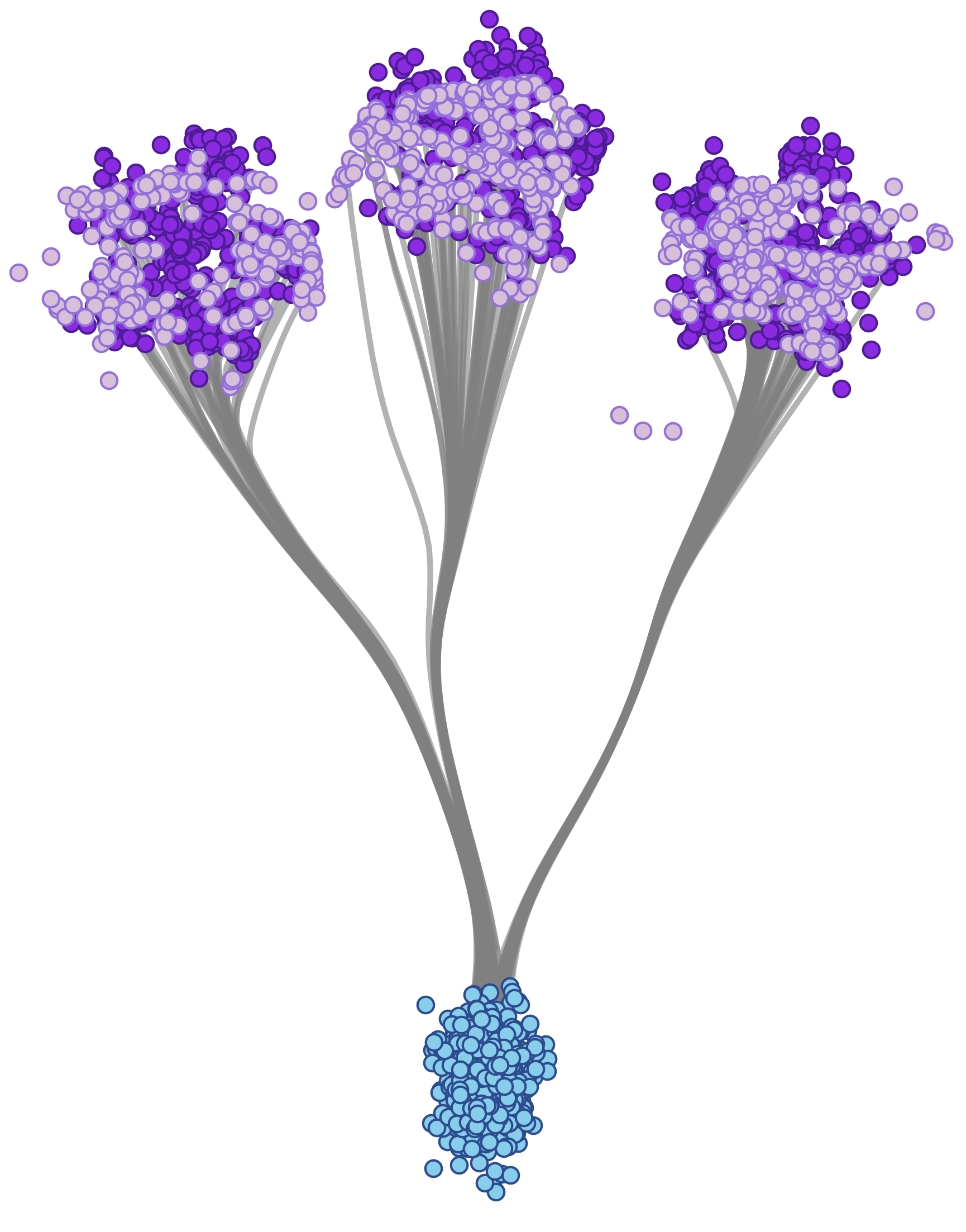}
    \caption{18  (BOTA)}
\end{subfigure}

\caption{BOTA Gaussian mixture results for increasing number of Gaussians. Top row: Flow Matching. Bottom row: BOTA ($\alpha = 0.5$).}
\label{fig:gaussian-mixtures-horizontal}
\vspace{-5mm}
\end{figure*}

\section{Experiments}
\vspace{-3mm}
\subsection{Gaussian Mixtures}
We compare how two flow-based methods transport mass from a single source distribution to a highly multi-modal target. The target is a mixture of $K$ clusters (“branches”) arranged at the top; the source is a cluster (or mix of two clusters) at the bottom. We train the same-size models with (i) standard FM and (ii) BOTA. Each curve shows a sample’s trajectory from source (blue) to its destination cluster (purple).
As $K$ grows (3 $\rightarrow$ 6 $\rightarrow$ 18), FM learns many independent, fan-out paths that crowd and cross, offering little shared routing. 

BOTA instead discovers a shared “trunk” that later splits into branches, yielding short, structured, tree-like transport. The qualitative gap widens with more branches. Both use models the same time-conditioned MLP $v_\theta(x, t)$: 64-$d$ time embedding $\rightarrow$ 3×256 SiLU layers $\rightarrow$ 2-D output; integrated with 10 fixed Euler steps over $t\in [0,1]$. Training is identical: Adam ($lr=10^{-3}$, batch 256, 10k iterations, seed 42. FM uses the standard flow-matching objective along linear source–target couplings. For the results, please see Figure \ref{fig:gaussian-mixtures-horizontal}. As you can notice, FM produces independent straight-line arms; BOTA discover shared trunks before splitting. 

\subsection{Biological data}
\begin{table}
    \centering
    \caption{Comparison of methods on the Tedsim dataset (50D PCA); mean $\pm$ std over 5 seeds.}
    \label{tab:comparison_Tedsim_extended}
    \small
    \begin{tabular}{lccccc}
        \toprule
        Metric & BSBM & CNF & CFM & FM & BOTA \\
        \midrule
        $W_1$ & $17.72 \pm 0.12$ & $13.76 \pm 0.10$ & $12.01 \pm 0.05$ & $12.03 \pm 0.13$ & $\mathbf{11.86 \pm 0.07}$ \\
        $W_2$ & $17.96 \pm 0.15$ & $13.81 \pm 0.11$ & $12.12 \pm 0.04$ & $12.15 \pm 0.16$ & $\mathbf{12.02 \pm 0.09}$ \\
        RBF-MMD & $0.63 \pm 0.007$ & $0.51 \pm 0.005$ & $0.11 \pm 0.002$ & $0.15 \pm 0.007$ & $\mathbf{0.11 \pm 0.003}$ \\
        \bottomrule
    \end{tabular}
\end{table}

\begin{wrapfigure}{r}{0.4\linewidth} %
    \centering
    \includegraphics[width=\linewidth]{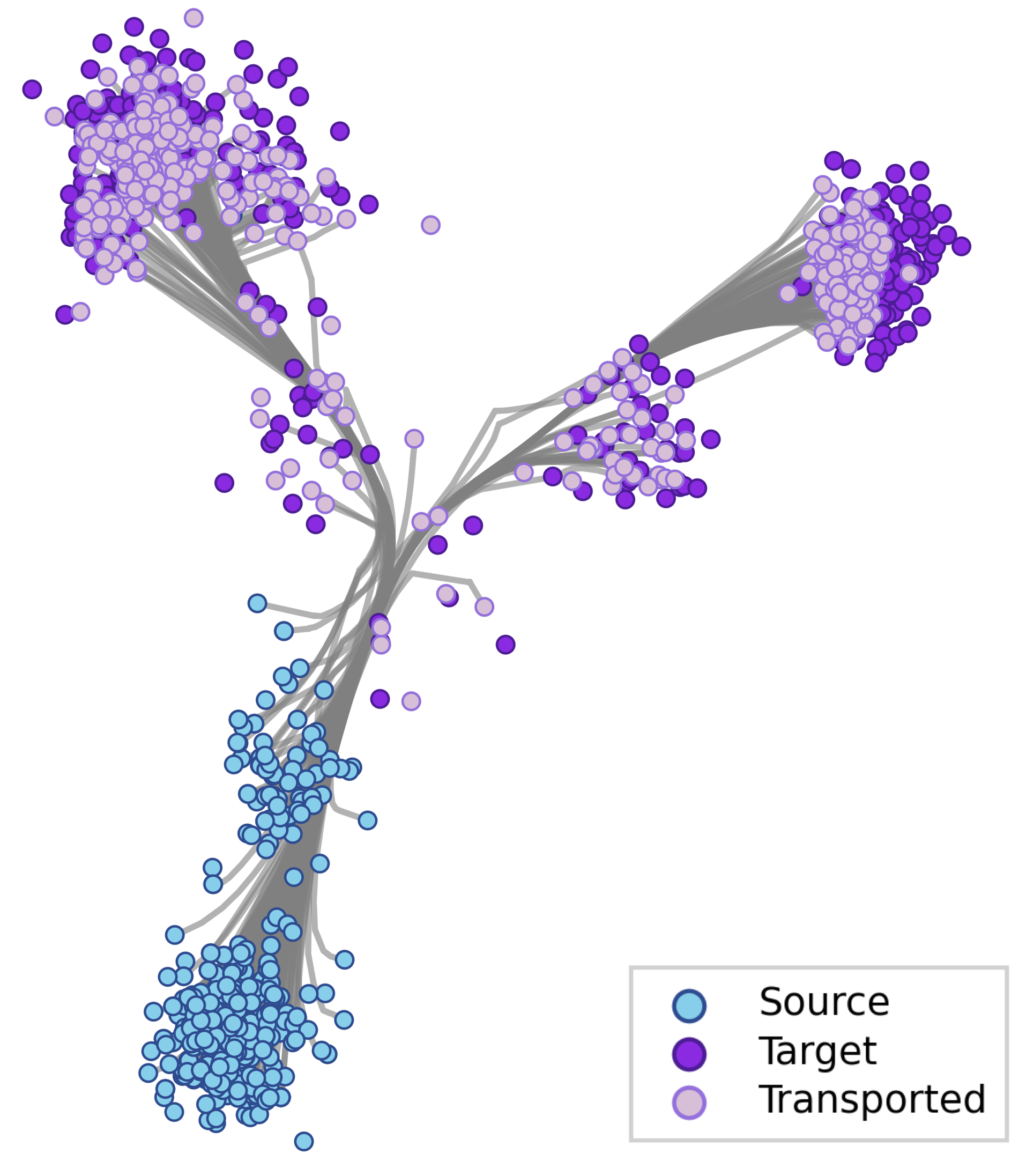}
    \caption{BOTA results on complicated biology Tedsim dataset with $\alpha = 0.5$.}
    \vspace{-5mm}
    \label{fig:tedsim}
\end{wrapfigure}
To further validate our method, we use the Tedsim dataset (\cite{pan2022tedsim}) as a controlled reference point with known ground truth dynamics. Tedsim simulates a cellular differentiation process by modeling cell division from a root cell, generating both gene expression profiles and heritable lineage barcodes. This provides a complete record of a simple, branching differentiation process.

We apply our method to a specific Tedsim scenario, modeling the transport from a single progenitor cell to two distinct terminal states. This controlled experiment allows us to quantitatively assess our method's ability to accurately reconstruct branching trajectories where the true pathways are known beforehand. Please see Figure \ref{fig:tedsim} and Table \ref{tab:comparison_Tedsim_extended}. We use the same architecture for fair comparison for all types of models except for Mean-Flows which requires additional input $r$ to encoding the start of sampling trajectory. 

\textbf{Comparison baselines}: For image data Flow Matching (\textbf{FM})\cite{lipman2022flow} and Mean Flows (\textbf{MFs})\cite{geng2025mean} Branched Schrödinger Bridge Matching (BSBM)\cite{tang2025branched} and CNF\cite{chen2018neural} are used as baselines in our work. All methods are trained with exactly the same amount of iteration 100k, learning rate 1e-4 and optimizer Adam. The experiments highlight our model's effectiveness in reconstructing branching trajectories.  Our method produces the best results among all the models compared. By leveraging the natural branching structure inherent in the data, our approach achieves the highest performance score while maintaining computational efficiency. The parameter $\alpha$ ablation please see in Appendix.
\vspace{-3mm}
\subsection{Image data}
\begin{wrapfigure}{l}{0.35\textwidth}
    \centering
    \vspace{-2mm}
    \includegraphics[width=\linewidth]{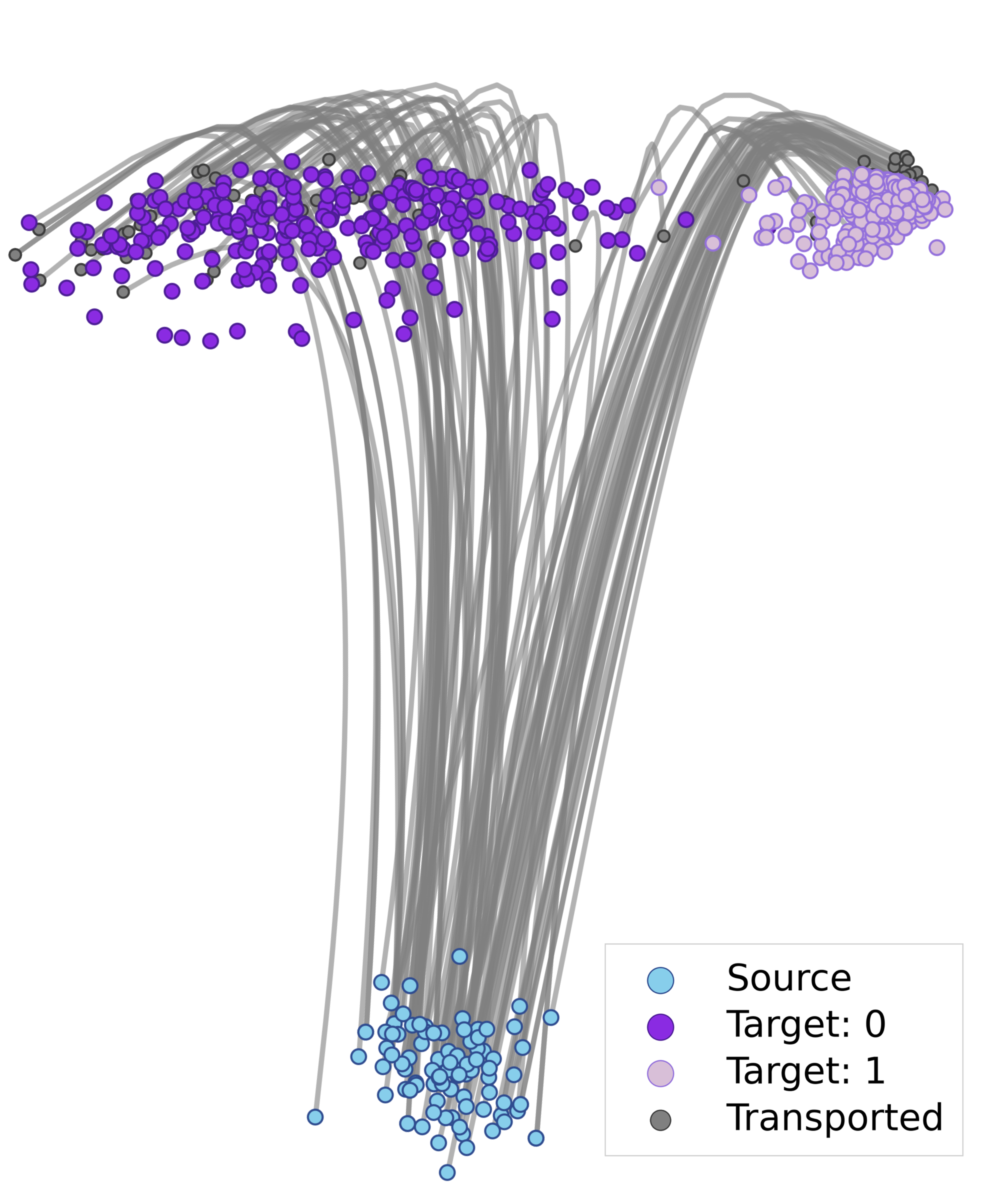}
    \caption{PCA of the learned flow for generating classes 0 and 1.}
    \label{fig:mnist_pca}
    \vspace{-4mm}
\end{wrapfigure}
\textbf{MNIST}. For our proof of concept with image data, we conducted experiments using the classic MNIST dataset \cite{deng2012mnist}. We designed two experiments: first, mapping digits into the two classes 0 and 1 to examine branching structures in the data, and second, mapping the entire dataset from the Gaussian distribution. As shown in Figure \ref{fig:mnist_pca}, our method successfully reveals branching structures in this image dataset as well. The hyperparameters setting was set equal to the one chosen for the biology experiment.

\begin{figure*}[h!]
    \centering
    \begin{subfigure}[b]{0.9\linewidth}
        \centering
        \includegraphics[width=\linewidth]{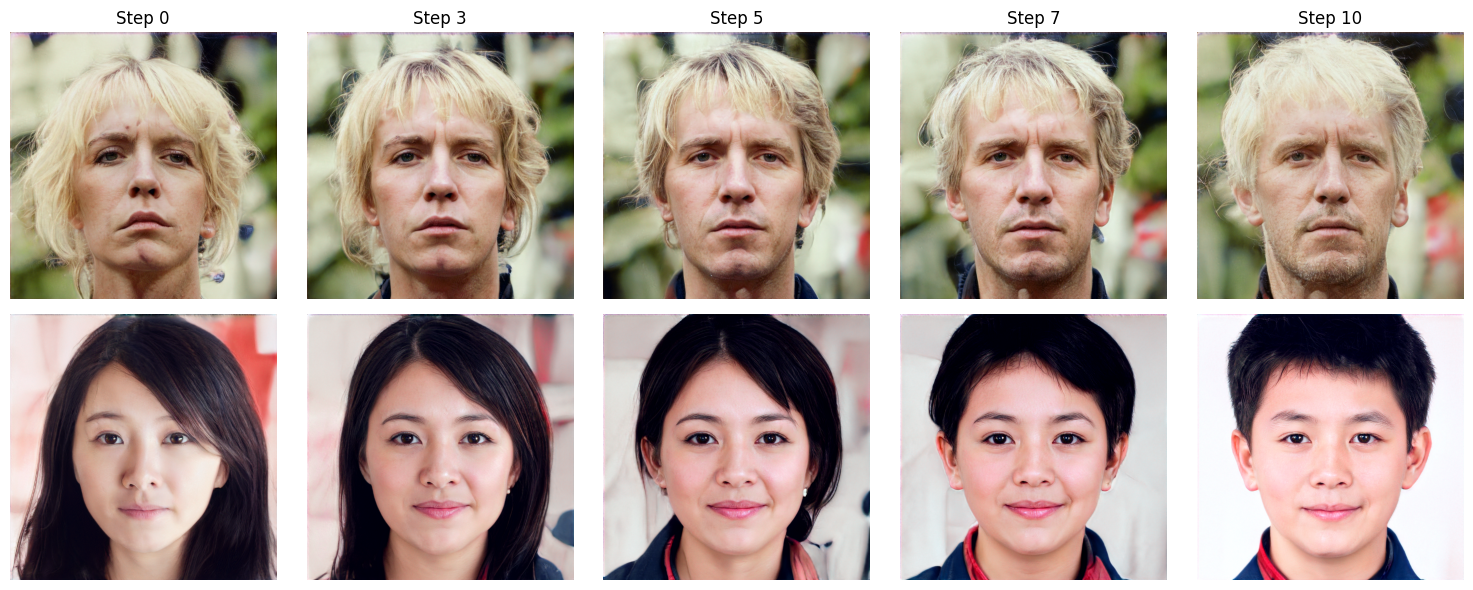}
    \end{subfigure}
    \caption{Branched flow interpolation on FFHQ dataset From left to right over the timestamps. Female $\rightarrow$ Male task.}
    \vspace{-3mm}
    \label{fig:ffhq}
\end{figure*}
\textbf{FFHQ}. All experiments are conducted in the latent space of a pretrained ALAE model on FFHQ 1024x1024 dataset. The ALAE is used \emph{only} for decoding latents to images during evaluation and remains frozen at all times. We load latent vectors paired with gender labels and split them into 60,000 training and 10,000 test samples. Let $\mathcal{D}$ denote the latent dimensionality inferred from the data. The goal of this experiment is to show that our method can produce a smooth and hierarchical interpolation between two classes. We learn a BOTA from source-class latents to target-class latents. (etc. female $\to$ male); mini-batches for $x_0$ and $x_1$ are sampled from the respective class-specific training subsets. We amortize model a time-dependent vector field $v_{\theta}(t,x)$ that defines the ODE $\frac{d}{dt}x(t) = v_{\theta}\bigl(t, x(t)\bigr)$  integrated from $t=0$ to $t=1$. 

\textbf{Architecture:} $v_\theta$ is an MLP with Tanh activations and hidden width 1024. A scalar time embedding (linear layer) is concatenated to $x$ before the MLP; the output is D-dimensional.  For the results please see Figure \ref{fig:ffhq}.
\vspace{-4mm}
\section{Related Work}
\vspace{-2mm}
\label{sec:related}
Branched optimal transport can be formalized in various ways, such as for traffic plans, irrigation flows, and branched networks. Some details on rectified flows, action matching, and generative flow networks are also provided. For example in \citep{maddalena2003variational}, with a single source supply $\mu$. The authors modeled the transportation network as a set of “fibers” $\chi(p, \cdot)$, where $\chi(p,t)$ represents the location of a particle $p \in \Omega$ in time $t$, and ($\Omega$ is an abstract probability space) \citep{bernot2005traffic}. Such approaches provide insight into how continuous mass might cluster around “branch nodes,” but they remain computationally intractable for high-dimensional data.

In parallel, the deep generative modeling community \cite{asadulaev2025shaped, tang2025branched} proposed Y-shaped and branched Generative Flows, a continuous-time flow framework that encourages branching of paths. However, these models do not explicitly solve the BOT optimization problem; they embed a BOT-inspired inductive bias in the model but do not guaranty a globally optimal branched transport plan for the given data. In other words, Y-shaped flows learn a hierarchy of transport heuristically within a single-stage network training, rather than computing or supervising with the ground-truth BOT.

\vspace{-2mm}
\section{Limitations and Broader Impact}
\vspace{-2mm}
Our method relies on a differentiable soft-atomic relaxation whose geometry depends on choices such as proxy atoms and kernel bandwidth may require task-specific tuning. The two-stage discrete optimization and neural amortization procedure also adds design complexity. This work may support applications involving structured branching processes, such as cellular differentiation and other hierarchical systems, and improves the expressivity of generative models. However, it inherits standard generative-modeling risks, including misuse for synthetic data generation and amplification of dataset biases, especially in sensitive domains.
\vspace{-2mm}
\section{Conclusion}
We introduce Branched Optimal Transport Amortization (BOTA), a scalable generative algorithm designed to overcome the inability of standard continuous-time models to capture hierarchical, tree-like data structures. By adapting the Benamou-Brenier formulation with a concave "economy of scale" cost function, BOTA encourages probability mass to aggregate along shared "trunks" before branching out to specific targets. To make this computationally feasible, the method utilizes a novel differentiable "soft-atomic" relaxation and amortizes the solution into a continuous neural vector field via flow matching. Experiments demonstrate that BOTA effectively recovers underlying branching geometries and outperforms baseline methods in reconstructing complex, multi-modal distributions.

\clearpage
\bibliographystyle{plainnat}
\bibliography{neurips_2026}

\clearpage
\section*{Extended Analysis}
\label{sec:extended_analysis}

\begin{equation}
\label{eq:soft-atomic-loss-appendix}
\mathcal{L}^\alpha(\theta)
= \int_{t_0}^{t_1} \sum_{k=1}^{K}
\underbrace{ \|\bar{v}_k(t)\|\, m_k(t)^{\alpha} }_{\text{\textcolor{Green}{soft branched cost}}} \, dt
- \lambda\, \underbrace{ \mathbb{E}_{x_{t_1} \sim \mu_1} \!\big[\log p_\theta(x_{t_1}, t_1)\big] }_{\textcolor{Blue}{\text{match data at } t_1}} .
\end{equation}

\textbf{Lemma 1 (Differentiable Approximation of the Instantaneous Branched Transport Cost).}
\textit{
Let $\hat{\rho}_t = \sum_{j=1}^B \rho_j \, \delta_{x_j(t)}$ be an empirical measure representing a particle approximation of a distribution at time $t$, with particle positions $\{x_j(t)\}_{j=1}^B \subset \mathbb{R}^d$ and velocities $\{v_j(t)\}_{j=1}^B \subset \mathbb{R}^d$.
Let $\{c_k\}_{k=1}^K$ be a set of fixed centers.
Define the \texttt{soft\_atomic\_cost} as
\[
C_{\text{soft}}(\{x_j\}, \{v_j\})
= \sum_{k=1}^K \|\bar{v}_k\| \, m_k^\alpha ,
\]
where for each center $c_k$,
\begin{align*}
m_k &= \sum_{j=1}^B w_{jk}\, \rho_j \quad &&(\text{Soft Mass}),\\
\bar{v}_k &= \frac{\sum_{j=1}^B w_{jk}\, v_j}{\sum_{j=1}^B w_{jk} + \varepsilon} \quad &&(\text{Soft Averaged Velocity}),
\end{align*}
with smooth weights
\[
w_{jk}
= \frac{\exp\!\left(-\|x_j - c_k\|^2 / 2\sigma^2\right)}
       {\sum_{\ell=1}^K \exp\!\left(-\|x_j - c_\ell\|^2 / 2\sigma^2\right)} .
\]
Then $C_{\text{soft}}$ is differentiable in $\{x_j, v_j\}$ for any $\sigma>0$, and as $\sigma \to 0$, it converges to the hard-assignment cost where each particle is assigned to its nearest center.
}

\begin{proof}
The instantaneous branched transport cost for a purely atomic measure
$\rho_t = \sum_i \rho_{t,i} \, \delta_{x_i}$ \citep{brasco2011benamou}
is given by
\[
\mathcal{F}(\rho_t) = \sum_i \|v_{t,i}\| \, \rho_{t,i}^\alpha .
\]
To approximate this functional with a differentiable surrogate, we discretize space into $K$ centers $\{c_k\}$ and replace the discontinuous Voronoi assignment with a soft weighting.

In the hard, non-differentiable case, the partition of space into Voronoi cells $\{V_k\}$ yields
\begin{align*}
m_k^{\text{hard}} &= \sum_{j=1}^B \rho_j \, \mathbf{1}_{x_j \in V_k}, \\
\bar{v}_k^{\text{hard}} &= \frac{\sum_{j=1}^B \rho_j \, v_j \, \mathbf{1}_{x_j \in V_k}}
{\sum_{j=1}^B \rho_j \, \mathbf{1}_{x_j \in V_k}} .
\end{align*}
This construction is non-differentiable because the indicator $\mathbf{1}_{x_j \in V_k}$ is discontinuous.

We therefore replace it with smooth, normalized kernel weights $w_{jk}$ as defined above.
The Gaussian kernel ensures that $w_{jk}$ is infinitely differentiable in $x_j$.
As $\sigma \to 0$, the kernel becomes sharply peaked, so that
\[
w_{jk} \;\to\;
\mathbf{1}\!\left[k = \arg\min_\ell \|x_j - c_\ell\|\right] .
\]
Consequently,
\[
m_k \;\to\; m_k^{\text{hard}},
\quad
\bar{v}_k \;\to\; \bar{v}_k^{\text{hard}} .
\]

Substituting these quantities into the definition of $C_{\text{soft}}$ gives
\[
C_{\text{soft}}
= \sum_{k=1}^K
\|\bar{v}_k\| \, m_k^{\alpha}
= \sum_{k=1}^K
\left\| \frac{\sum_j w_{jk} v_j}{\sum_j w_{jk}} \right\|
\left( \sum_j w_{jk} \rho_j \right)^\alpha .
\]
Each operation involved (sums, norms, powers, and exponentials) is differentiable for $\sigma > 0$ and $\varepsilon > 0$,
ensuring that $C_{\text{soft}}$ is a smooth, differentiable approximation of the branched transport cost.

In the limit $\sigma \to 0$,
the soft assignments converge to hard nearest-center assignments, and hence
\[
\lim_{\sigma \to 0} C_{\text{soft}}
= \sum_{k=1}^K
\|\bar{v}_k^{\text{hard}}\| \, (m_k^{\text{hard}})^\alpha .
\]
If the centers coincide with the atom positions ($c_k = x_k$) and $K = B$,
this expression recovers exactly
\[
\mathcal{F}(\rho_t) = \sum_{i=1}^B \|v_i\| \, \rho_i^\alpha .
\]
Thus, the proposed $C_{\text{soft}}$ provides a differentiable surrogate that converges to the atomic branched transport cost in the small-$\sigma$ limit.
\end{proof}

This differentiable reformulation replaces the hard, non-differentiable atomic constraint with a smooth objective that still promotes the desired clustering and branching behavior.
It enables gradient-based learning of velocity fields that generate transport maps with the geometric structure characteristic of branched optimal transport, thereby providing a practical and effective objective for solving the branched optimal transport problem.

\begin{figure*}[t]
\centering

\begin{subfigure}[b]{0.32\linewidth}
    \centering
    \includegraphics[height=0.22\textheight]{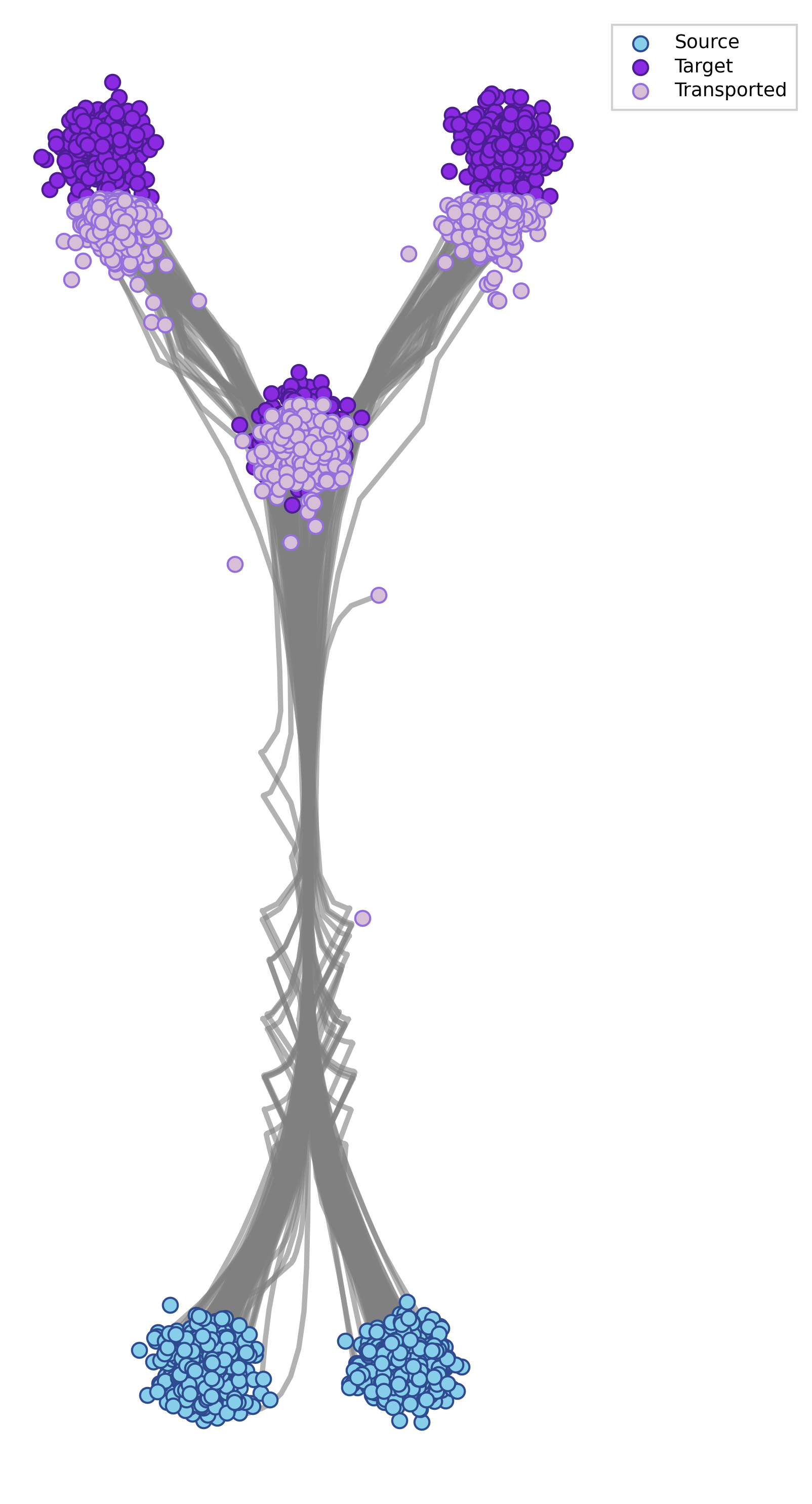}
    \caption{3 Gaussians ($\alpha=0.1$)}
\end{subfigure}
\hfill
\begin{subfigure}[b]{0.32\linewidth}
    \centering
    \includegraphics[height=0.22\textheight]{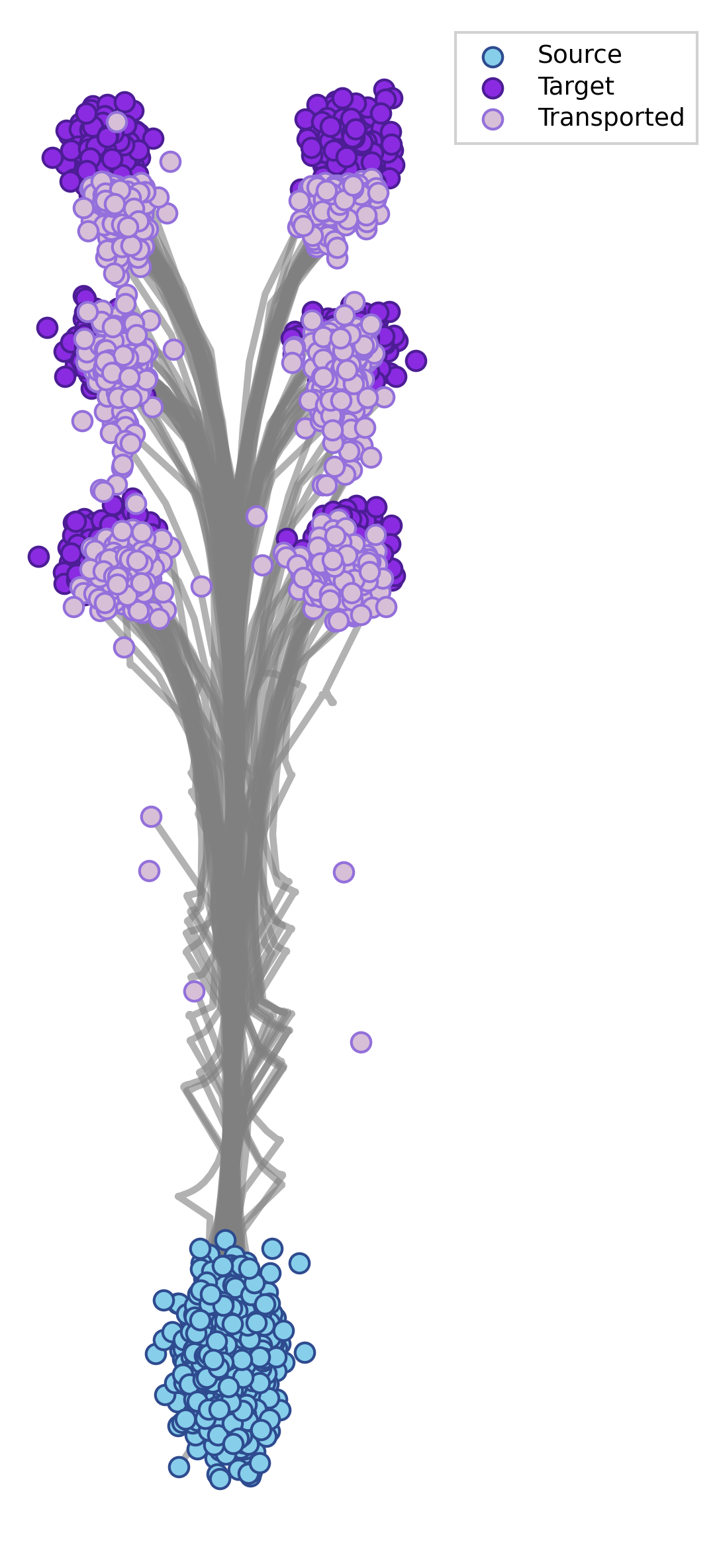}
    \caption{6 Gaussians ($\alpha=0.1$)}
\end{subfigure}
\hfill
\begin{subfigure}[b]{0.32\linewidth}
    \centering
    \includegraphics[height=0.22\textheight]{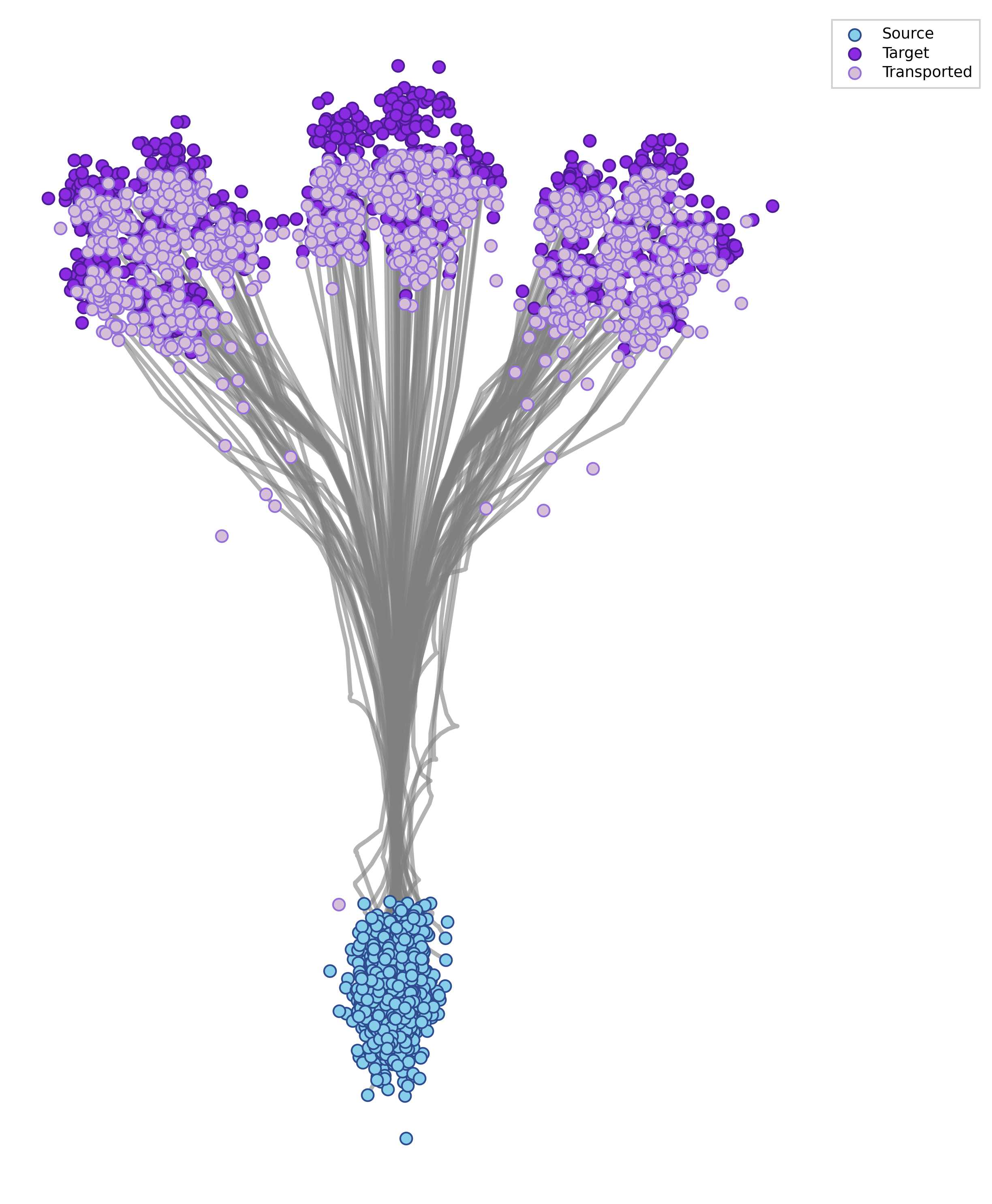}
    \caption{18 Gaussians ($\alpha=0.1$)}
\end{subfigure}

\vspace{3mm}

\begin{subfigure}[b]{0.32\linewidth}
    \centering
    \includegraphics[height=0.22\textheight]{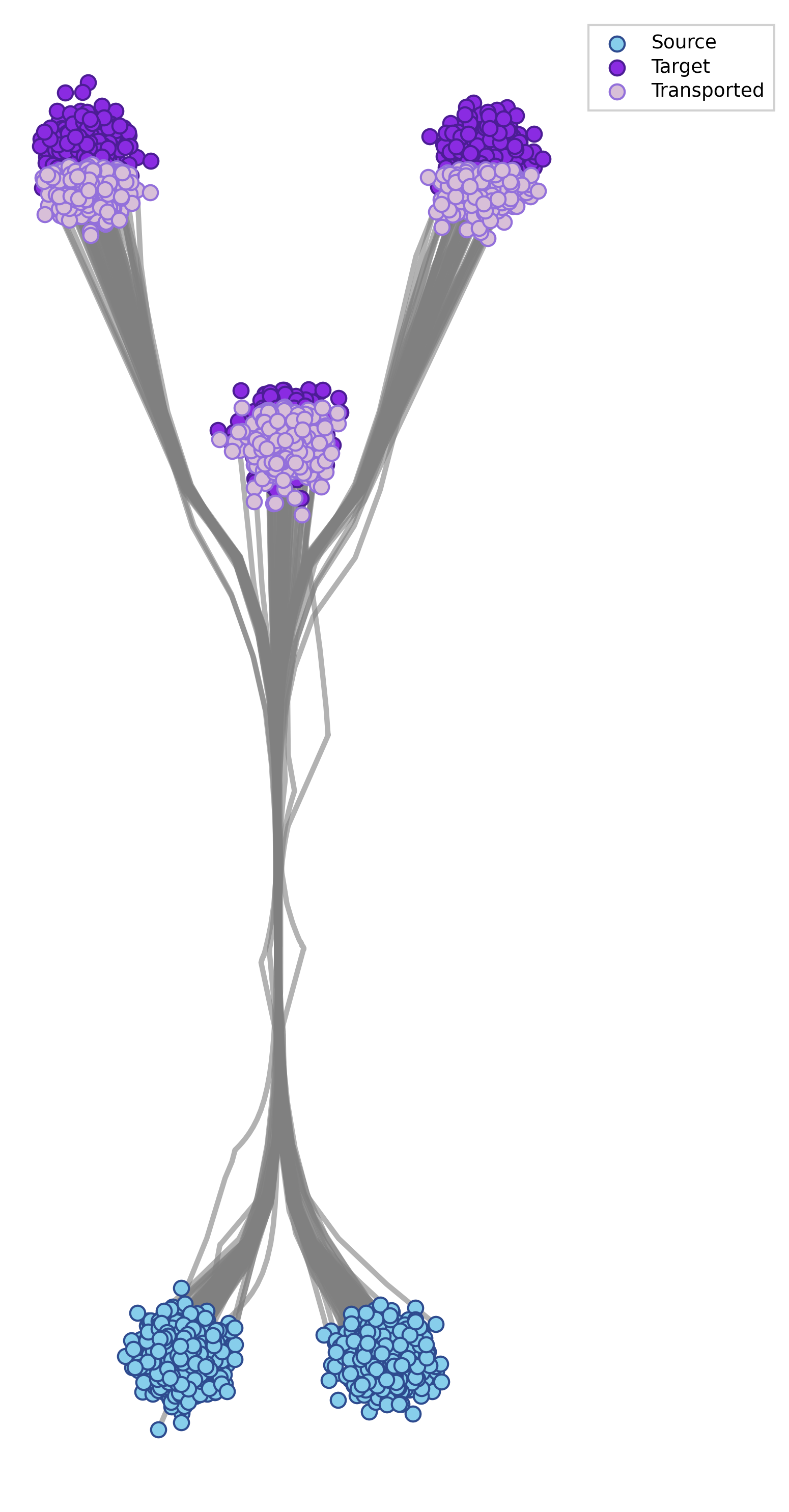}
    \caption{3 Gaussians ($\alpha=0.5$)}
\end{subfigure}
\hfill
\begin{subfigure}[b]{0.32\linewidth}
    \centering
    \includegraphics[height=0.22\textheight]{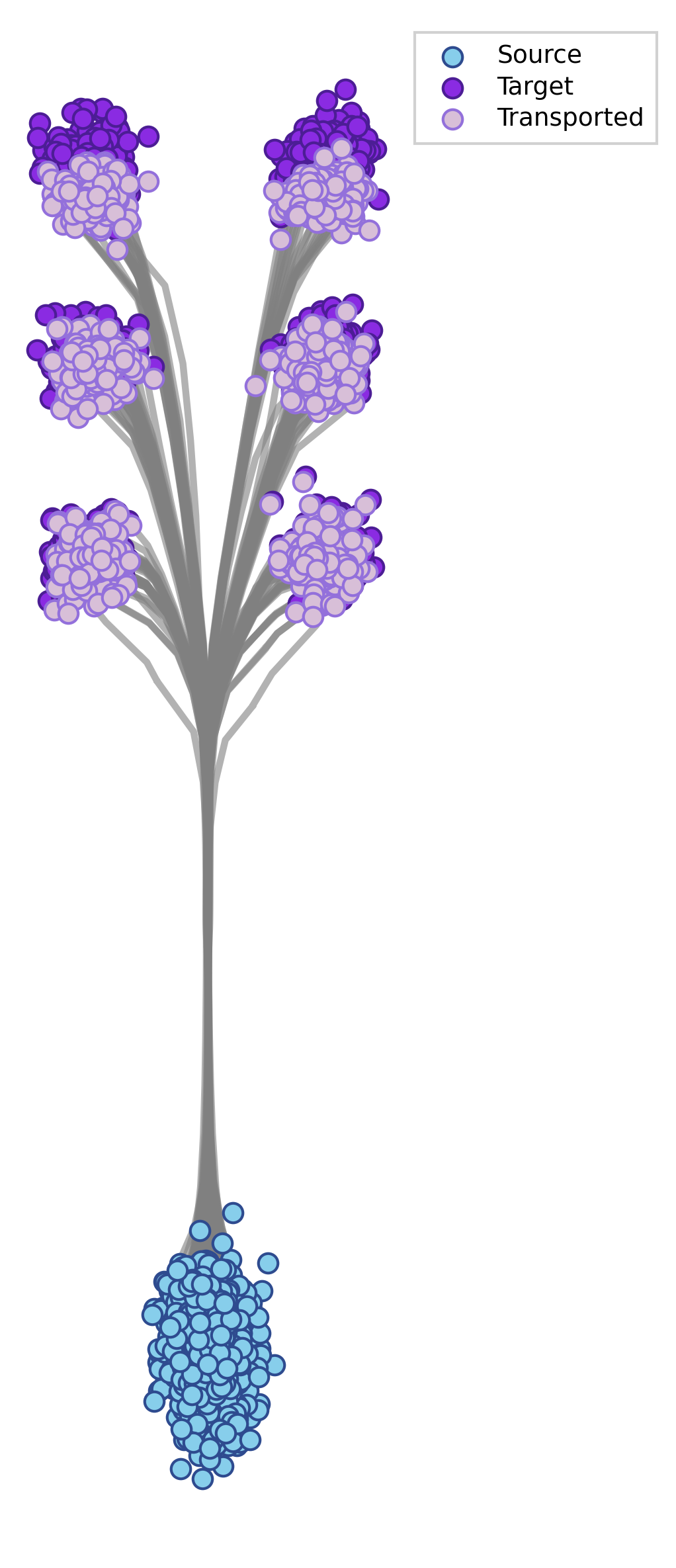}
    \caption{6 Gaussians ($\alpha=0.5$)}
\end{subfigure}
\hfill
\begin{subfigure}[b]{0.32\linewidth}
    \centering
    \includegraphics[height=0.22\textheight]{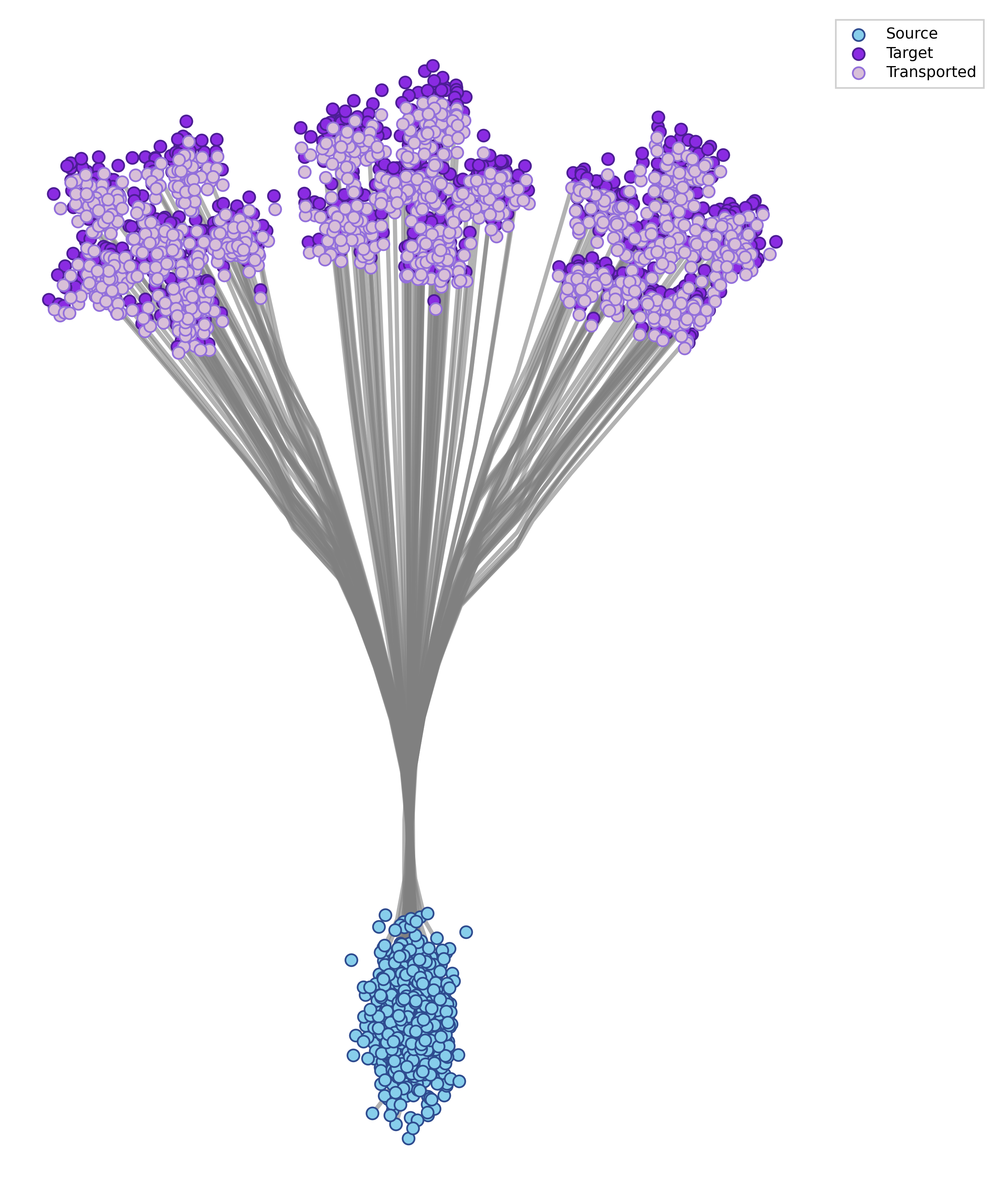}
    \caption{18 Gaussians ($\alpha=0.5$)}
\end{subfigure}

\begin{subfigure}[b]{0.32\linewidth}
    \centering
    \includegraphics[height=0.22\textheight]{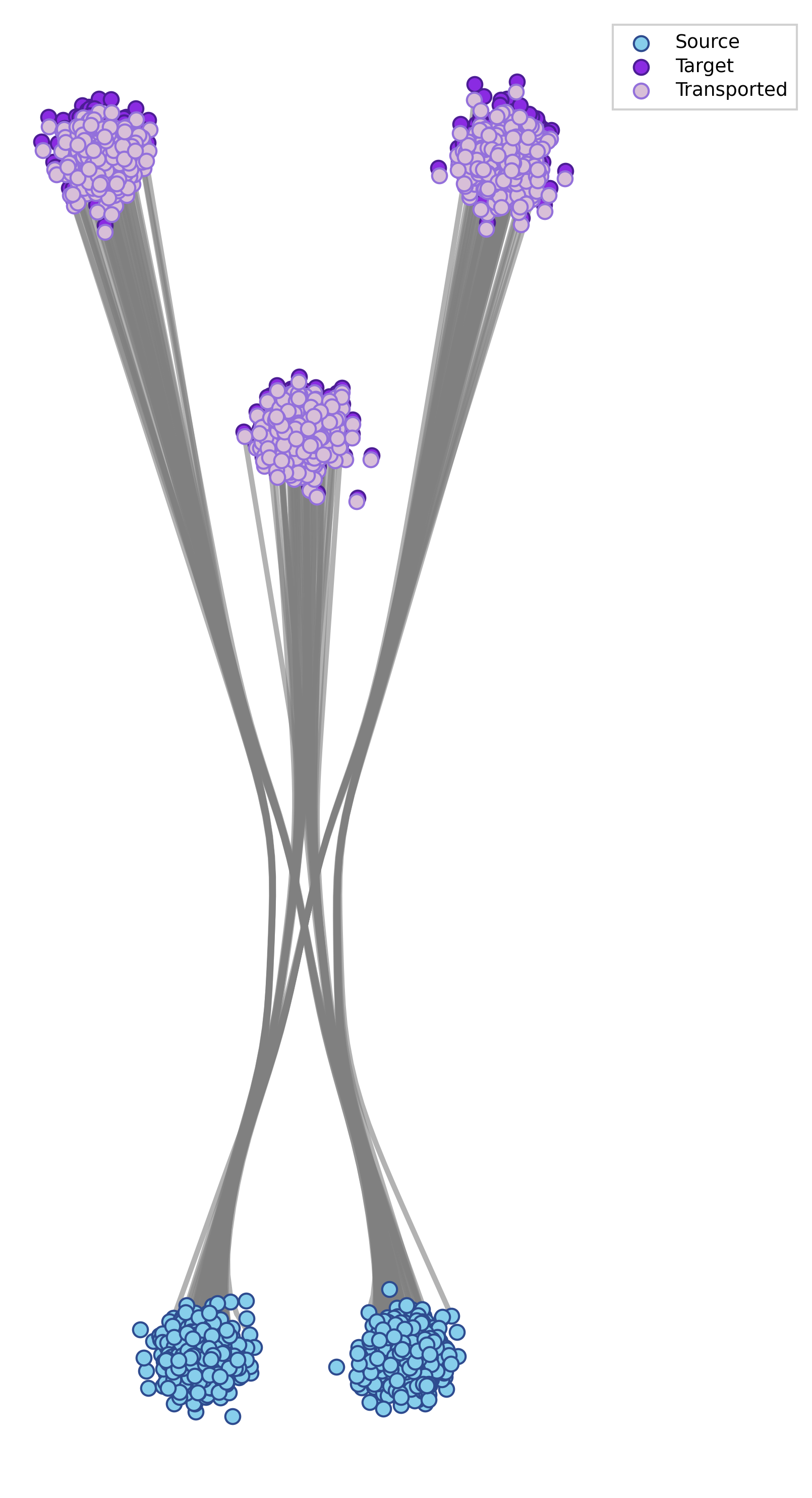}
    \caption{3 Gaussians ($\alpha=0.9$)}
\end{subfigure}
\hfill
\begin{subfigure}[b]{0.32\linewidth}
    \centering
    \includegraphics[height=0.22\textheight]{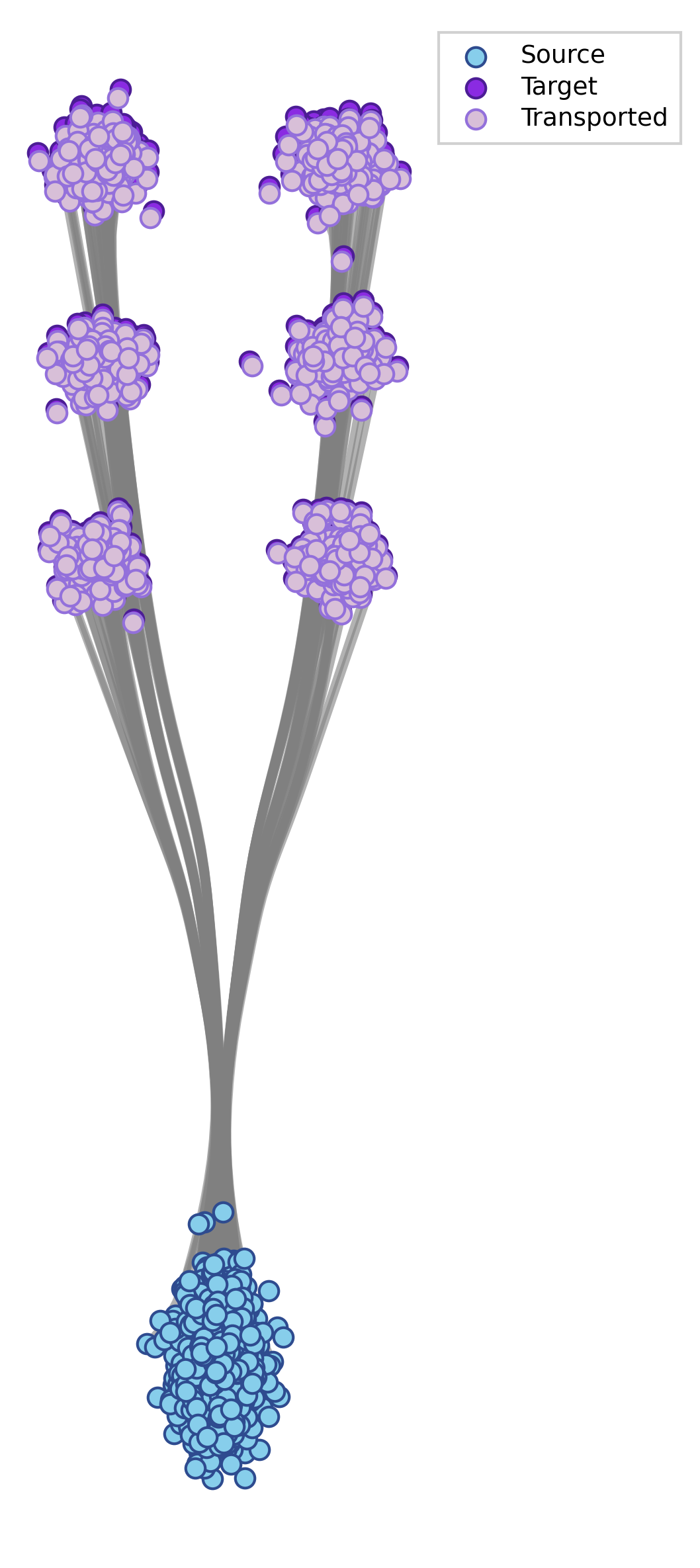}
    \caption{6 Gaussians ($\alpha=0.9$)}
\end{subfigure}
\hfill
\begin{subfigure}[b]{0.32\linewidth}
    \centering
    \includegraphics[height=0.22\textheight]{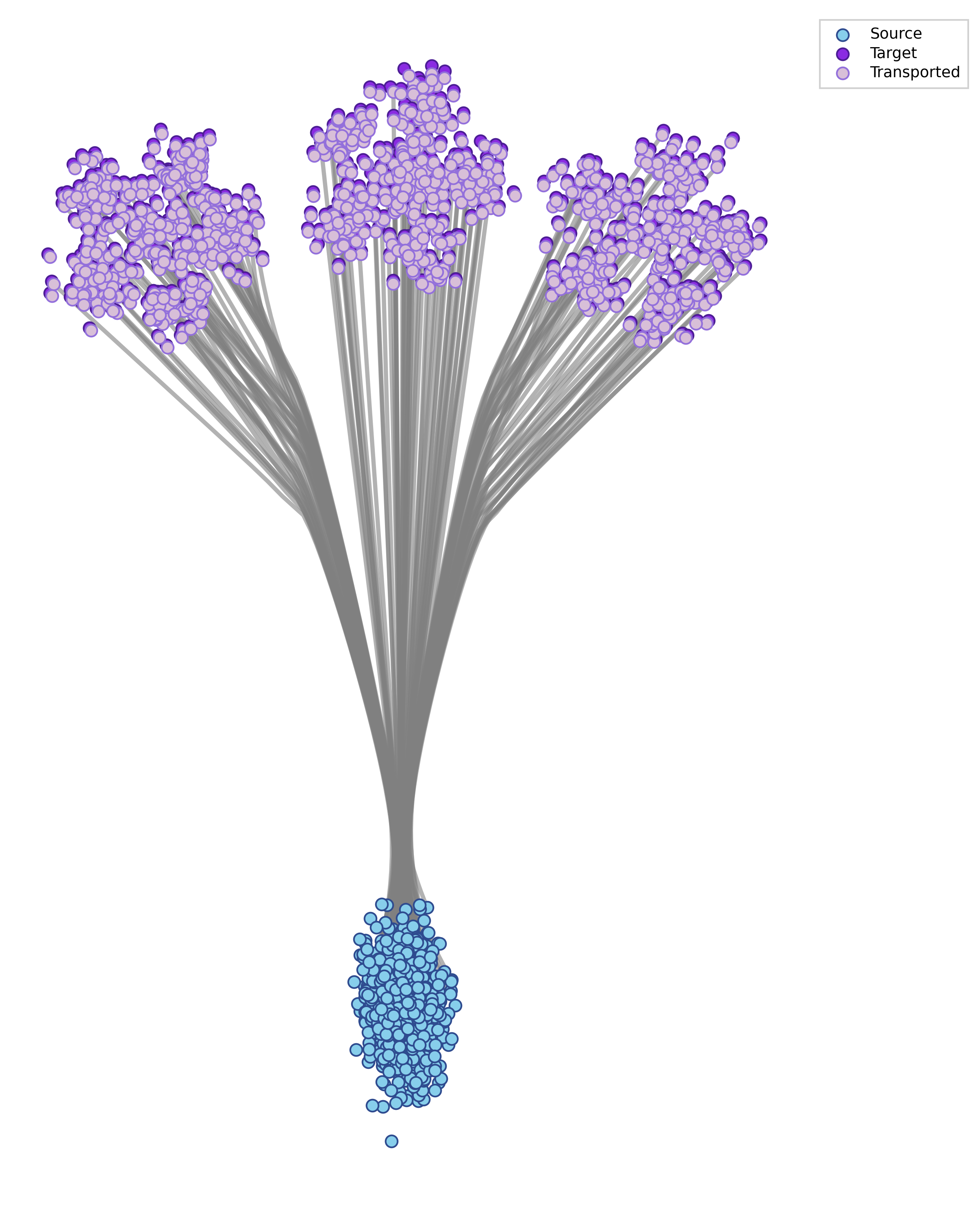}
    \caption{18 Gaussians ($\alpha=0.9$)}
\end{subfigure}

\caption{BOTA Gaussian mixture results for increasing number of Gaussians (3, 6, 18) and different branching parameters $\alpha$. Top row: $\alpha = 0.1$. Middle row: $\alpha = 0.5$. Bottom row: $\alpha = 0.9$.}
\label{fig:alpha-ablations}
\end{figure*}

\subsection*{Gaussian Data}

In this section, we further analyze the role of the branching parameter $\alpha$ in shaping transport geometry through a series of Gaussian ablations (Fig.~\ref{fig:alpha-ablations}). These experiments provide both qualitative and structural insight into how trajectory behavior evolves across regimes. For larger values of $\alpha$, trajectories become increasingly independent, closely resembling classical optimal transport solutions. In contrast, smaller values of $\alpha$ encourage mass to follow shared pathways, giving rise to pronounced trunk-and-branch structures. This transition highlights how $\alpha$ governs the balance between independent transport and cooperative routing, ultimately controlling the geometric organization of mass flow.

\subsection*{Image Data}

\begin{wrapfigure}{r}{0.50\linewidth}
\centering

\includegraphics[width=0.46\linewidth]{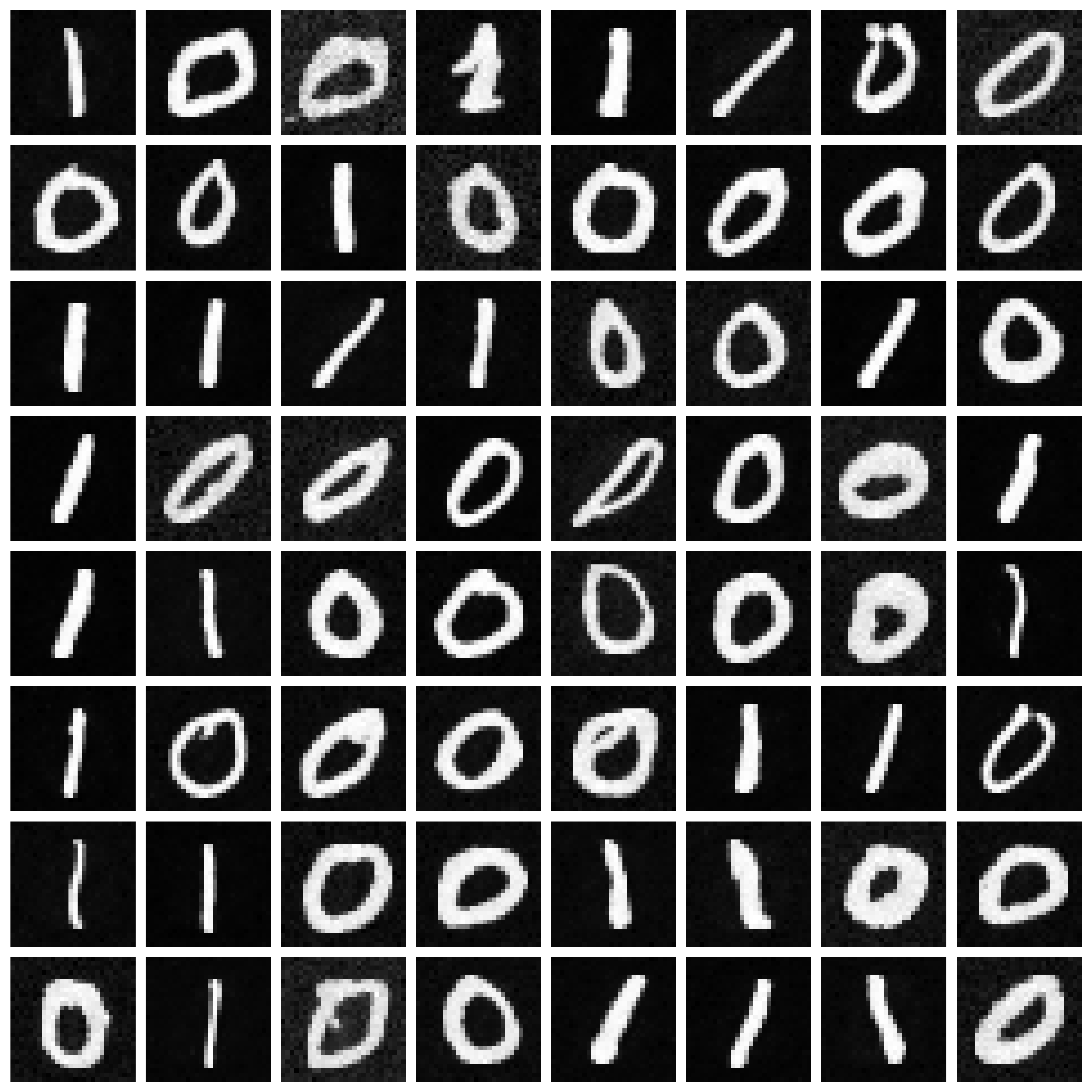}
\hfill
\includegraphics[width=0.51\linewidth]{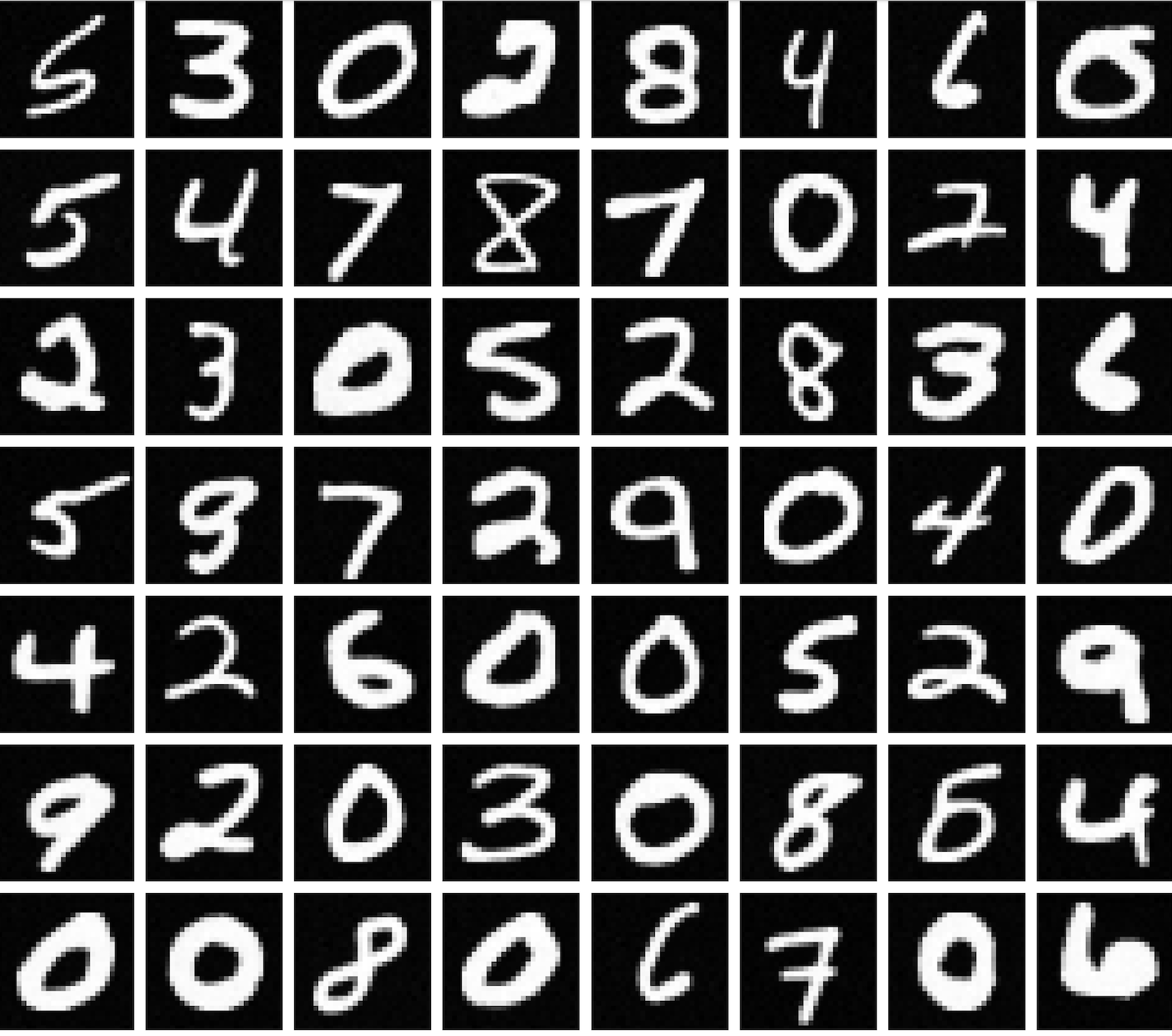}

\vspace{3mm}
\captionof{figure}{Examples of MNIST digits generated by BOTA.}
\label{fig:mnist_bota}

\end{wrapfigure}

In this supplementary section, we extend the analysis presented in the main paper with additional qualitative and quantitative results on MNIST and FFHQ.

For MNIST, Figure~\ref{fig:mnist_bota} provides additional samples generated by BOTA.  
These results illustrate that the model produces clean, well-formed digits across classes, with consistent stroke thickness and minimal visual artifacts.  
This complements the main paper by showing that BOTA maintains stable trajectories even in discrete, multidimensional datasets, resulting in reliable class-conditional generations.

\begin{table}[h!]
    \centering
    \caption{FID scores comparison on the FFHQ dataset.}
    \label{tab:fid_scores}
    \begin{tabular}{l c}
        \hline
        \textbf{Method} & \textbf{FID} \\
        \hline
        FM   & $10.61 \pm 1.39$ \\
        BOTA & $\mathbf{10.46 \pm 1.80}$ \\
        \hline
    \end{tabular}
\end{table}
For high-resolution natural images, Table~\ref{tab:fid_scores} reports the full FID scores on FFHQ.  
Also BOTA achieves a lower FID compared to Flow Matching, indicating that the learned transport field aligns more accurately with the underlying data distribution.  
The improvement in FID is consistent across seeds and corroborates the qualitative differences shown in the main paper.

\subsection*{Biological Data}

We evaluate BOTA on the Tedsim dataset, a controlled benchmark with known branching differentiation dynamics. Tedsim simulates cellular development from a single progenitor into multiple terminal states, providing both gene expression data and lineage information, making it well-suited for assessing the recovery of branching trajectories.

\begin{figure}[t]
\centering

\begin{subfigure}{0.48\linewidth}
\centering
\begin{minipage}[c][0.38\textheight][c]{\linewidth} %
\centering
\includegraphics[width=\linewidth]{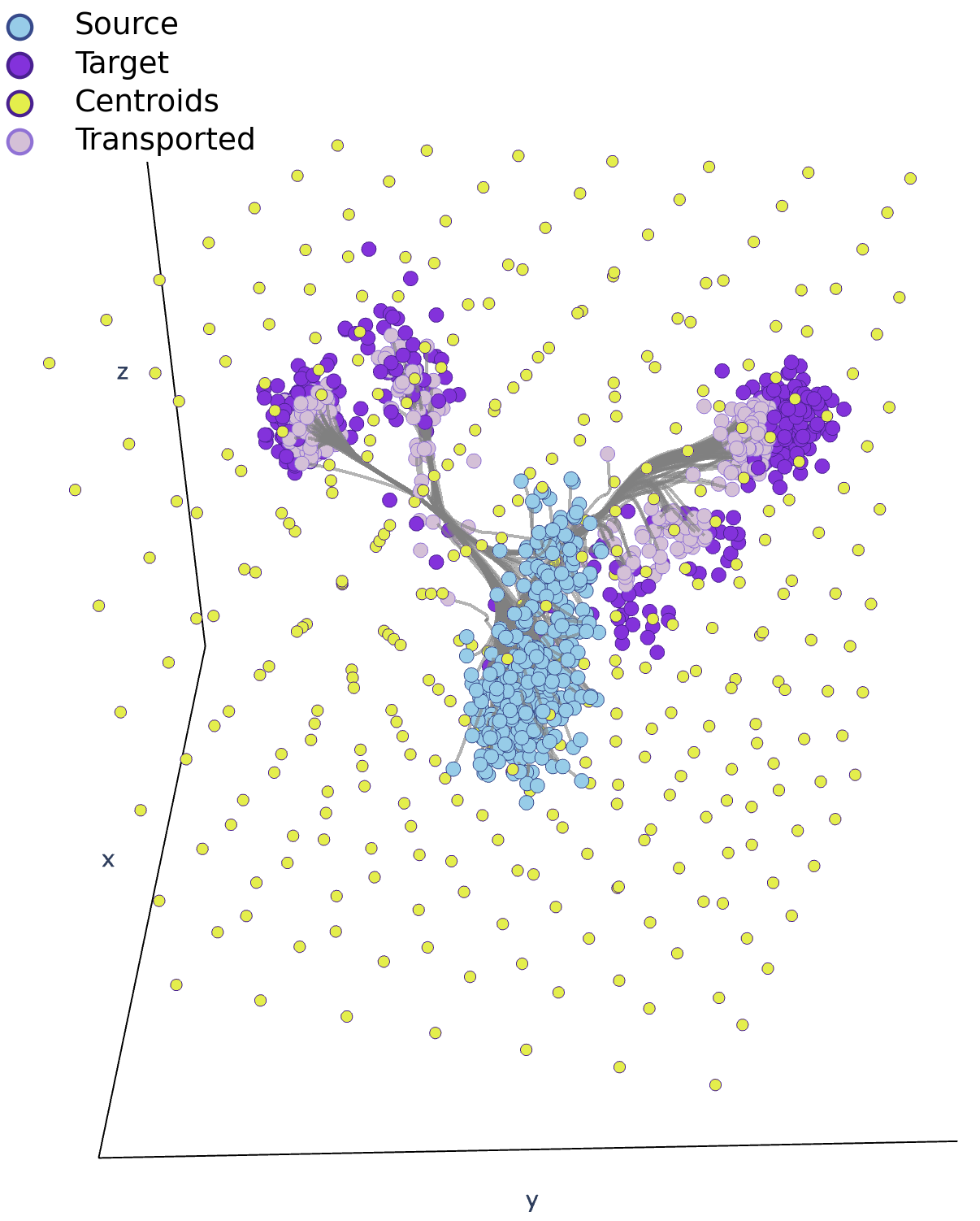}
\end{minipage}
\caption{Learned branched transport on Tedsim dataset.}
\end{subfigure}
\hfill
\begin{subfigure}{0.48\linewidth}
\centering
\begin{minipage}[c][0.38\textheight][c]{\linewidth} %
\centering

\renewcommand{\arraystretch}{1.2}
\resizebox{0.85\linewidth}{!}{
\begin{tabular}{l c}
\hline
\textbf{Method} & \textbf{Cost $\downarrow$} \\
\hline
FM   & $405.08 \pm 1.00$ \\
CFM  & $408.88 \pm 1.00$ \\
BOTA & $\mathbf{174.81 \pm 5.02}$ \\
\hline
\end{tabular}
}

\end{minipage}
\caption{Soft-atomic cost comparison.}
\end{subfigure}

\vspace{3mm}

\begin{subfigure}{\linewidth}
\centering
\includegraphics[width=\linewidth]{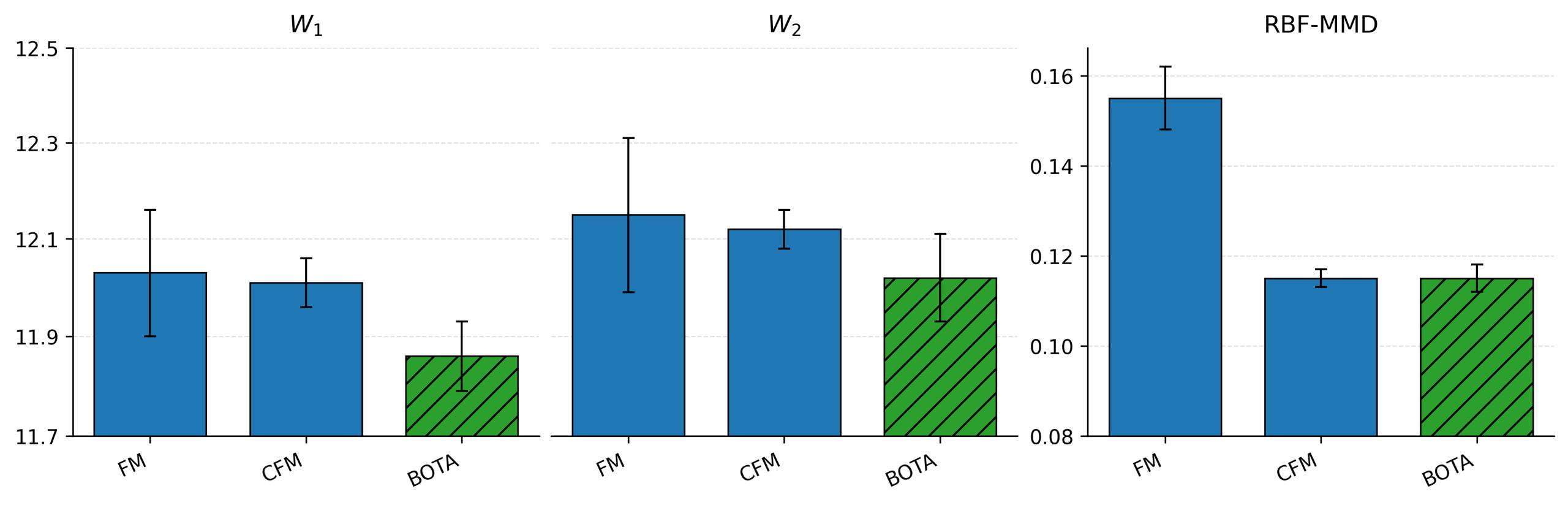}
\caption{Quantitative metrics ($W_1$, $W_2$, RBF-MMD) with $\alpha = 0.5$.}
\end{subfigure}

\caption{
BOTA results on the Tedsim dataset combining qualitative transport structure and quantitative evaluation.
}
\label{fig:tedsim_all}

\end{figure}

As shown in Fig.~\ref{fig:tedsim_all}, BOTA accurately reconstructs the underlying bifurcation structure, producing smooth trajectories that align with the true developmental paths. The model captures a shared progenitor trunk followed by coherent lineage splits, without mode collapse or spurious branching. Quantitatively, BOTA achieves substantially lower soft-atomic cost and competitive distributional metrics ($W_1$, $W_2$, RBF-MMD), confirming that the learned transport is both structurally meaningful and statistically consistent.

\clearpage
\paragraph{Runtime and complexity.}
The branched machinery adds a bounded, one-off training overhead and no sampling overhead. Stage 1 optimizes the discrete velocity tensor $V \in \mathbb{R}^{T \times N \times D}$ directly, with no neural network: each iteration costs $O(T \cdot N \cdot K \cdot D)$, linear in the number of particles $N$, with no $N^2$ term (unlike an OT or Sinkhorn pairing, which costs $O(N^2 D)$ before training begins). Stage 2 is unmodified flow matching, so its cost equals that of the FM baseline. Table~\ref{tab:runtime} reports wall-clock training times on Tedsim (50-D PCA) under the identical training protocol of Table~\ref{tab:comparison_Tedsim_extended}. In total, BOTA trains in roughly $2\times$ the time of FM, paid once offline, while sampling cost is identical to FM since inference integrates a single velocity field of the same architecture.

\begin{table}[h]
    \centering
    \caption{Training cost on Tedsim (50-D PCA). The discrete solver stage involves no network training; the amortization stage is standard flow matching. Sampling cost is identical for all methods.}
    \label{tab:runtime}
    \small
    \begin{tabular}{lccc}
        \toprule
         & FM & CFM & BOTA \\
        \midrule
        Solver stage & --- & --- & 425 s (no network) \\
        Network training & 360 s & 480 s & 360 s (unmodified FM) \\
        Total training & $1\times$ FM & $1.3\times$ FM & $\approx 2\times$ FM \\
        Per-iteration cost in $N$ & $O(N)$ & $O(N^2)$ if OT-paired & $O(T \cdot N \cdot K \cdot D)$, linear in $N$ \\
        Sampling cost & $1\times$ & $1\times$ & $1\times$ \\
        \bottomrule
    \end{tabular}
\end{table}

\end{document}